%% file: main.tex
\documentclass[10pt,twocolumn,letterpaper]{article}
\usepackage[pagenumbers]{cvpr}
\input{preamble}

\definecolor{cvprblue}{rgb}{0.21,0.49,0.74}
\usepackage[pagebackref,breaklinks,colorlinks,allcolors=cvprblue]{hyperref}
\def\confName{CVPR}
\def\confYear{2027}

\title{Aligning Physical Reasoning and Video Generation via \\  Coupled Reinforcement Learning}

\usepackage[cvprbody]{sjtudenglab}

\title{\huge Aligning Physical Reasoning and Video Generation\\
    via Coupled Reinforcement Learning}
\author[1,2,*]{Yanzhe Hu}
\author[1,*]{Xiang Li}
\author[3]{Yuzheng Zhang}
\author[4]{Qiuyue Li}
\author[1,\dagger]{Zhijie Deng}
\affiliation[1]{\,\mbox{Shanghai Jiao Tong University}}
\affiliation[2]{\,\mbox{Huazhong University of Science and Technology}}
\affiliation[3]{\,\mbox{East China Normal University}}
\affiliation[4]{\,\mbox{The Hong Kong Polytechnic University}}
\contribution[*]{\,Equal contribution}
\contribution[\dagger]{\,Corresponding author}
\abstract{\input{sections/abstract}}
\metadata[Correspondence]{\sffamily Zhijie Deng$^{\dagger}$\enspace
    \href{mailto:zhijied@sjtu.edu.cn}{zhijied@sjtu.edu.cn}}
\date{\sffamily November 25, 2026}

\hypersetup{
    pdftitle={Aligning Physical Reasoning and Video Generation via Coupled Reinforcement Learning},
    pdfauthor={Yanzhe Hu, Xiang Li, Yuzheng Zhang, Qiuyue Li, Zhijie Deng}
}
\AtBeginDocument{
    \fancyheadoffset[R]{-0.3cm}
    \fancyhead[R]{\raisebox{9pt}{\includegraphics[height=35pt]{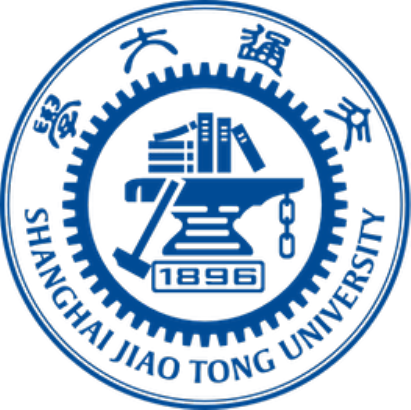}}}
}

\begin{document}
\begin{metacover}
\maketitle
\enlargethispage{12pt}

\begingroup
\setlength{\intextsep}{5pt}
\begin{figure}[H]
    \centering
    \captionsetup{skip=5pt}
    \includegraphics[width=\linewidth]{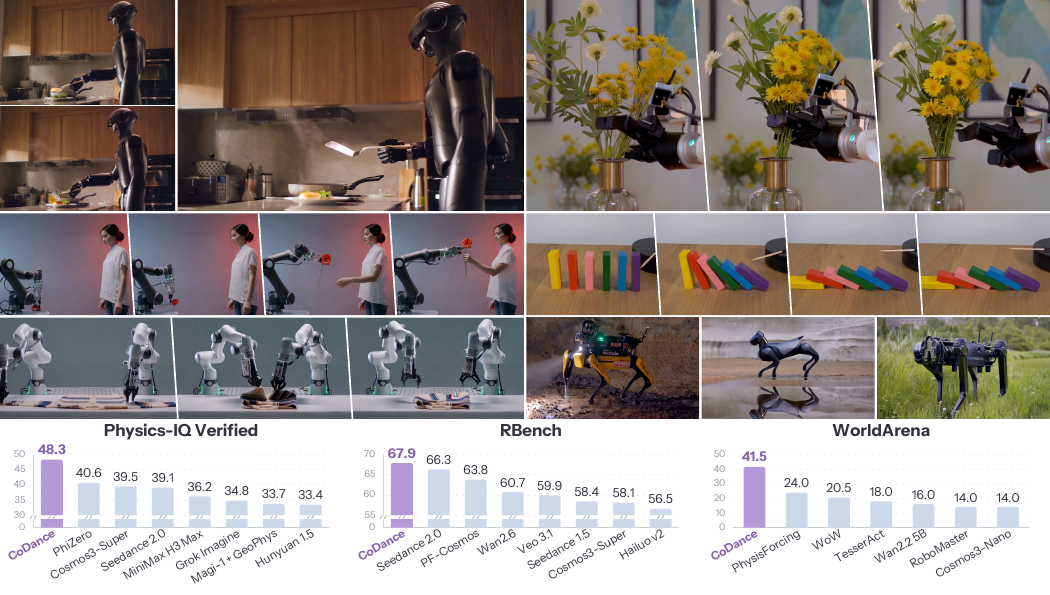}
    \caption{\textbf{CoDance generations and benchmark results.} Top: selected frames from CoDance videos of manipulation, locomotion, and domino interactions. Bottom: IQ-Score on Physics-IQ Verified~\citep{DBLP:journals/corr/abs-2606-18943}, average score on RBench~\citep{DBLP:journals/corr/abs-2601-15282}, and mean closed-loop success rate (\%) on two WorldArena~\citep{DBLP:journals/aiopen/ShangLMSJWJZTSGWLSZCZTC26} action planning tasks (\textit{adjust bottle}, \textit{click bell}). CoDance leads the compared models on all three benchmarks, demonstrating physically consistent video generation, embodied interaction modeling, and downstream action planning.}
    \label{fig:teaser}
\end{figure}
\endgroup

\end{metacover}

\fancyfoot[C]{\thepage}
\twocolumn[{%
  \begin{minipage}{\textwidth}
    \captionsetup{type=figure}%
    \centering
    \includegraphics[width=\linewidth]{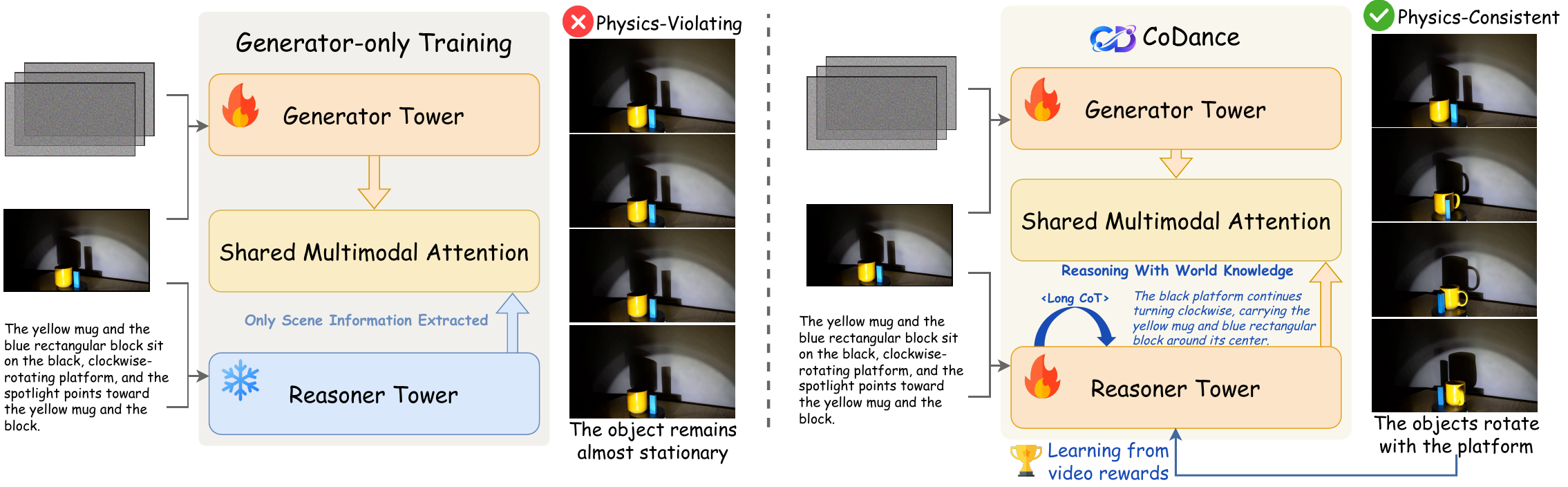}
    \caption{\textbf{Generator-only training versus CoDance.} In this example, generator-only training with a frozen reasoner leaves the objects nearly stationary on a rotating platform (left), whereas CoDance jointly trains the reasoner and generator with video-level feedback to produce the corresponding object motion (right). Flame and snowflake denote trainable and frozen towers.}
    \label{fig:overview}
  \end{minipage}%
}]

\input{sections/body}

{\small
\bibliographystyle{cvprunsrt}
\bibliography{references}
}
\label{paper:submission-end}

\input{sections/appendix}

\end{document}

%% file: preamble.tex
\input{math_commands.tex}

\usepackage{algorithm}
\usepackage{algpseudocode}
\usepackage{colortbl}
\usepackage{multirow}
\usepackage{pifont}
\usepackage{tabularx}
\usepackage{microtype}

\definecolor{categoryblue}{RGB}{238,246,252}
\definecolor{ourspurple}{RGB}{245,240,255}
\newcommand{\cmark}{\ding{51}}
\newcommand{\xmark}{\ding{55}}
\graphicspath{{figure/}}

\makeatletter
\@ifpackageloaded{lineno}{%
  \IfFormatAtLeastTF{2025-06-01}{%
    \@ifpackagelater{lineno}{2026/03/08}{}{%
      \AtBeginDocument{%
        \RemoveFromHook{build/column/before}[lineno]%
        \def\@LN@makecol{%
          \@LN@orig@makecol
          \setbox\@outputbox\vbox{%
            \boxmaxdepth\@maxdepth
            \protected@write\@auxout{}{%
              \string\def\string\@LN@column{\if@firstcolumn1\else2\fi}%
            }%
            \box\@outputbox
          }%
        }%
      }%
    }%
  }{}%
}{}%
\makeatother



%% file: math_commands.tex
\usepackage{amsmath,amsfonts,bm}

\def\eqref#1{equation~\ref{#1}}

\def\1{\bm{1}}

\DeclareMathAlphabet{\mathsfit}{\encodingdefault}{\sfdefault}{m}{sl}
\SetMathAlphabet{\mathsfit}{bold}{\encodingdefault}{\sfdefault}{bx}{n}



%% file: sections/abstract.tex
Image-to-video models can produce visually realistic videos, which, yet, often violate the physical constraints of motion and interactions.
This is harmful for the downstream embodied prediction and planning.
Recent remedies focus on reasoning-based prompt refinement or post-training of the video generator, which are, arguably, sub-optimal.
This paper proposes to jointly optimizing the reasoning and generation parts of the image-to-video models to exhaustively infer physical dynamics and realize them in the generated videos.
The resultant \textbf{CoDance} framework accomplish the joint optimization in a unified reinforcement learning procedure, dubbed \textbf{CoDanceRL}.
Concretely, CoDanceRL couples the updates of a vision-language reasoner and a flow-based video generator through shared physical feedback, combining Coupled Policy Optimization, Physics-Aware Feedback, and Hierarchical Reward Normalization.
As a result, CoDance achieves an IQ-Score of \textbf{48.3} on Physics-IQ Verified. On RBench, it achieves an average embodied video generation score of \textbf{67.9}. For closed-loop action planning with the video model frozen, CoDance achieves an average success rate of \textbf{41.5\%} across two \mbox{WorldArena} tasks. CoDance outperforms all compared models across these evaluations, demonstrating improved physical consistency and the utility for downstream action planning.

%% file: sections/body.tex
\section{Introduction}
\label{sec:introduction} 
Recent advances in image-to-video (I2V) generation have enabled increasingly realistic visual continuations of an observed scene~\citep{DBLP:journals/corr/abs-2604-14148,DBLP:journals/corr/abs-2606-02800}. Beyond content creation, generated videos can represent candidate futures for embodied prediction and planning~\mbox{\citep{DBLP:conf/nips/DuY0DN0SA23,DBLP:journals/corr/abs-2508-00795}}. These applications require physically plausible motion and interactions, yet visually realistic videos can still violate physical constraints~\citep{DBLP:conf/wacv/MotamedCSJG26}, limiting their utility in embodied settings.

Some models have begun to integrate physical understanding with video generation~\citep{DBLP:journals/corr/abs-2606-02800,DBLP:journals/corr/abs-2603-24458,DBLP:conf/iclr/WeiLYWWWGC26}. These models unify vision-language models (VLMs) and diffusion models within a single architecture. The VLM serves as both a reasoner and a conditioning encoder, while the diffusion model serves as the generator. In this architecture, the VLM provides input representations and draws on its world knowledge and reasoning capabilities to formulate physical guidance, which the generator translates into concrete motion and interactions.

From this perspective, existing approaches to improving physical consistency fall into two broad categories. One line of work refines prompts through physical reasoning~\citep{DBLP:journals/corr/abs-2511-20280,DBLP:journals/corr/abs-2603-03505} or trains a VLM as a prompt rewriter~\citep{DBLP:conf/eccv/LuoLWZTWGF26}, providing more explicit guidance to a fixed generator. Another post-trains the generator using physics-informed supervision~\citep{DBLP:journals/corr/abs-2606-28128} or rewards~\citep{DBLP:conf/eccv/CaiLJWSLCJWTDJYH26,DBLP:journals/corr/abs-2512-00425}. However, the benefit of better guidance depends on whether the fixed generator can realize the corresponding dynamics. When the reasoner also serves as the conditioning encoder, updating it can alter the conditioning representations, potentially creating a mismatch with the fixed generator. Conversely, optimizing the generator alone can improve its response to existing conditioning, but does not optimize how the VLM uses its world knowledge and reasoning to guide generation.

These limitations lead us to a key insight: the reasoner and generator should adapt together, with the physical consistency of the resulting videos providing a shared learning signal. Through this process, the reasoner learns to provide physical guidance that better supports generation, while the generator adapts to the updated guidance and conditioning representations and improves its ability to realize the intended motion and interactions. 

Building on this insight, we introduce \textbf{CoDance}, an I2V model built on Cosmos3-Nano~\citep{DBLP:journals/corr/abs-2606-02800} for physically consistent video generation. Given the initial image and user prompt, its vision-language reasoner infers the scene's physical dynamics and rewrites the prompt accordingly, with the resulting representations conditioning the flow-based video generator through shared multimodal attention, as shown in Figure~\ref{fig:overview}.
To develop these capabilities, we train CoDance with \textbf{CoDanceRL}, a coupled reinforcement-learning framework with three components. \textit{Coupled Policy Optimization} alternates updates to the reasoner and generator using shared feedback from generated videos. \textit{Physics-Aware Feedback} compares generated videos with real references through complementary physical-fidelity and geometric trajectory rewards. \textit{Hierarchical Reward Normalization} constructs prompt- and video-level advantages by separately normalizing these rewards within the nested rollout hierarchy.

Empirically, CoDance improves physical consistency and embodied video generation, with gains extending to downstream closed-loop action planning (Figure~\ref{fig:teaser}). On Physics-IQ Verified~\citep{DBLP:journals/corr/abs-2606-18943}, it raises the IQ-Score from 34.8 to 48.3 relative to Cosmos3-Nano, with improvements also observed under a held-out VLM judge excluded from reward optimization. On RBench~\citep{DBLP:journals/corr/abs-2601-15282}, CoDance achieves an average score of 67.9, surpassing Seedance 2.0 (66.3). For downstream planning, it increases the average success rate across two WorldArena~\citep{DBLP:journals/aiopen/ShangLMSJWJZTSGWLSZCZTC26} tasks from 14.0\% to 41.5\% relative to the base model. CoDance leads all compared models on the aggregate metrics of these three evaluations, demonstrating improved physical consistency and planning utility.

\section{Preliminaries} 
\label{sec:preliminary}
\textbf{Flow-based image-to-video generation.}
Image-to-video models generate videos from reference images~\citep{DBLP:journals/corr/abs-2311-04145,DBLP:journals/corr/abs-2311-15127}.
In this work, we consider a model conditioned on both a reference image $I$
and a text prompt $p$, with $c=(I,p)$.
Let $z_0$ denote a clean video latent and
$\xi \sim \mathcal{N}(0,\mathbf{I})$ denote Gaussian noise.
Under a linear flow-matching path~\citep{DBLP:conf/iclr/LipmanCBNL23},
the intermediate latent is defined as
$z_t = (1-t)z_0 + t\xi,t \in [0,1].$
A DiT~\citep{DBLP:conf/iccv/PeeblesX23} parameterizes a conditional velocity field
$v_\theta(z_t,t,c)$, which is trained to predict the
target velocity $\xi-z_0$ by minimizing the expected squared error
\begin{equation}
    \mathcal{L}_{\mathrm{FM}}(\theta)
    =
    \mathbb{E}_{z_0,c,t,\xi}
    \left[
        \left\|
            v_\theta(z_t,t,c)-(\xi-z_0)
        \right\|_2^2
    \right].
\end{equation}
Generation starts from Gaussian noise at $t=1$ and
integrates the learned velocity field toward $t=0$,
after which the resulting latent is decoded into a video~\citep{DBLP:conf/iclr/LipmanCBNL23}.

\textbf{Response-Level GRPO.}
We use a response-level variant of Group Relative Policy
Optimization (GRPO)~\citep{DBLP:journals/corr/abs-2402-03300},
which estimates advantages from groups of sampled outputs.
Given a condition $c$, a group of prompt rewrites
$\{p_i\}_{i=1}^{G}$ is sampled from the old policy
$\pi_{\phi_{\mathrm{old}}}(\cdot \mid c)$.
For reward scores $\{R_i\}_{i=1}^{G}$, the group-relative
advantage is
$\hat{A}_i=(R_i-\mu_R)/(\sigma_R+\epsilon_{\mathrm{norm}})$,
where $\mu_R$ and $\sigma_R$ are the group mean and
standard deviation, and $\epsilon_{\mathrm{norm}}>0$ ensures numerical
stability.
The policy is optimized by maximizing the following objective
\begin{equation}
\label{eq:grpo}
\begin{aligned}
\mathcal{J}_{\mathrm{GRPO}}(\phi)
&=\mathbb{E}_{\{p_i\}_{i=1}^{G}\sim\pi_{\phi_{\mathrm{old}}}}
\\[-2pt]
&\quad \Bigg[
\tfrac{1}{G}\sum_{i=1}^{G}
\min\!\Big(\rho_i(\phi)\hat{A}_i,
\\[-2pt]
&\qquad\quad
\mathrm{clip}\!\left(\rho_i(\phi),1-\epsilon,1+\epsilon\right)\hat{A}_i\Big)
\\[-2pt]
&\qquad-\beta D_{\mathrm{KL}}\!\left[\pi_\phi\,\|\,\pi_{\phi_{\mathrm{ref}}}\right]
\Bigg],
\end{aligned}
\end{equation}
where
$\rho_i(\phi)=
\frac{\pi_\phi(p_i\mid c)}
{\pi_{\phi_{\mathrm{old}}}(p_i\mid c)}$
is the probability ratio for the full response, and
$\pi_{\phi_{\mathrm{ref}}}$ is the fixed reference policy.
Here, $\epsilon$ is the clipping threshold and
$\beta$ is the KL coefficient.

\textbf{DanceGRPO for Flow Matching Diffusion Models.}
DanceGRPO~\citep{DBLP:journals/corr/abs-2505-07818} extends GRPO to visual
generation by treating the stochastic sampling process as a
sequence of policy transitions. For a flow-based video model,
we discretize the reverse-time dynamics with a stochastic
sampler and denote the transition density at sampling step
$n$ for the $j$-th generated video by
$p_\theta(z_{j,n-1}\mid z_{j,n},c)$, where $c$ is the
generation condition. Here, $n=N,\ldots,1$ indexes the $N$
discrete sampling steps from Gaussian noise $z_{j,N}=\xi_j$
to the final latent $z_{j,0}$.
The corresponding probability ratio is
$\rho_{j,n}(\theta)
=
\frac{
    p_\theta(z_{j,n-1}\mid z_{j,n},c)
}{
    p_{\theta_{\mathrm{old}}}
    (z_{j,n-1}\mid z_{j,n},c)
}$.
Final video rewards yield group-relative advantages.
DanceGRPO averages the clipped surrogate in
Eq.~\ref{eq:grpo} over videos and steps, omitting the KL regularization term from the generator objective.

\begin{figure*}[t]
    \centering
    \includegraphics[width=\linewidth]{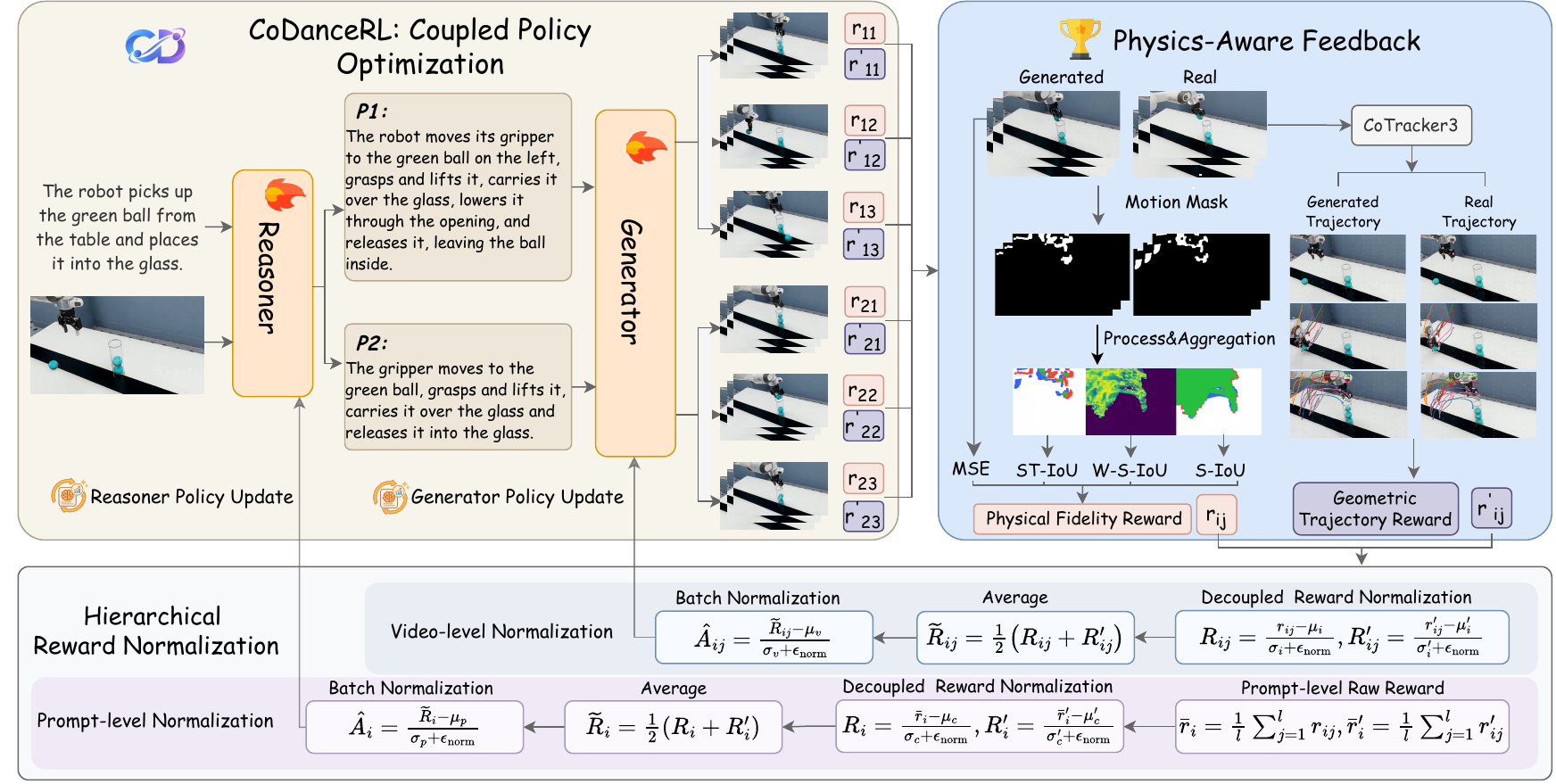}
    \caption{\textbf{Overview of CoDanceRL.} The reasoner samples prompt rewrites, and the generator produces multiple video rollouts for each rewrite. Complementary physical-fidelity and trajectory rewards are hierarchically normalized into video- and prompt-level advantages that update the generator and reasoner, respectively, within the same training iteration using the shared video feedback.}
    \label{fig:codancerl-framework}
\end{figure*}

\section{CoDanceRL for Physically Consistent Video Generation}
\label{sec:method}
CoDanceRL jointly adapts the reasoner's physical guidance and the generator's ability to realize it. As shown in Figure~\ref{fig:codancerl-framework}, a nested rollout samples multiple rewrites and multiple videos per rewrite. Reference-based physical rewards then supply two learning signals: the relative quality of videos under each rewrite, and the average quality achieved by each rewrite. These signals update both policies throughout the full duration of the same coupled training process.

\subsection{Coupled Policy Optimization}
\label{sec:optimization}
The reasoner should learn which rewrites lead to better videos under the current generator, and the generator should learn to improve its realizations of those rewrites. CoDanceRL couples these decisions through a shared nested rollout. Both policies are initialized by supervised fine-tuning (SFT; Supplementary Section~\ref{app:sft-details}) and remain trainable throughout RL. Each training example consists of an input $c=(I,p)$ and its associated real continuation video $V^\star$.

\paragraph{Nested rollout.}
We fix the policies at snapshots $\phi_{\mathrm{old}}$ and $\theta_{\mathrm{old}}$ during collection. The reasoner samples $G$ prompt rewrites $\{p_i\}_{i=1}^{G}\sim\pi_{\phi_{\mathrm{old}}}(\cdot\mid c)$. For $c_i=(I,p_i)$, let $h_i(\phi)=E_\phi(c_i)$ denote the language-pathway conditioning features used by joint attention. The encoder shares the reasoner's parameters $\phi$, so changing $\phi$ can change both the sampled rewrite and its conditioning features.

For each rewrite, the generator samples $l$ latent trajectories conditioned on $I$ and $h_i(\phi_{\mathrm{old}})$. Starting from Gaussian noise $z_{ij,N}=\xi_{ij}$, it samples $z_{ij,n-1}\sim p_{\theta_{\mathrm{old}}}(\cdot\mid z_{ij,n},I,h_i(\phi_{\mathrm{old}}))$ for $n=N,\ldots,1$. Decoding $z_{ij,0}$ gives $V_{ij}$, yielding $G l$ videos per input. The rewards in Section~\ref{sec:rewards} score each generated video against the real reference $V^\star$.

\paragraph{Policy-specific updates.}
Hierarchical normalization (Section~\ref{sec:normalization}) derives $\hat A_i$ from the mean raw rewards of the $l$ videos under $p_i$, and $\hat A_{ij}$ from comparisons within that rewrite. The first signal evaluates the guidance provided by a rewrite; the second evaluates its individual realizations. Let $\mathcal S_\epsilon(\rho,A)$ denote the clipped surrogate inside the sum in Eq.~\ref{eq:grpo}. We use distinct clipping thresholds $\epsilon_{\mathrm R}$ and $\epsilon_{\mathrm G}$; $\mathcal S_{\mathrm R}$ additionally applies the reasoner's dual clipping.

Using the response-level ratio $\rho_i(\phi)$ from Section~\ref{sec:preliminary}, the reasoner maximizes the following policy objective
\begin{equation}
\label{eq:reasoner_objective}
\begin{aligned}
\mathcal{J}_{\mathrm{R}}(\phi)
&=\mathbb{E}\Bigg[
\frac{1}{G}\sum_{i=1}^{G}
\mathcal S_{\mathrm R}\!\left(\rho_i(\phi),\hat A_i\right)
\\[-2pt]
&\quad-\beta D_{\mathrm{KL}}\!\left[
\pi_\phi(\cdot\mid c)\,\|\,
\pi_{\phi_{\mathrm{ref}}}(\cdot\mid c)
\right]
\Bigg].
\end{aligned}
\end{equation}

For the generator, we make the encoder dependence explicit in the generator's sampling transition ratio:
\begin{equation}
\label{eq:coupled_generator_ratio}
\rho_{ij,n}(\theta;\phi)=
\frac{p_\theta(z_{ij,n-1}\mid z_{ij,n},I,h_i(\phi))}
{p_{\theta_{\mathrm{old}}}(z_{ij,n-1}\mid z_{ij,n},I,h_i(\phi_{\mathrm{old}}))}.
\end{equation}
The numerator uses the reasoner's pre-update encoder state, held fixed when differentiating with respect to $\theta$; the denominator is the rollout transition density. All sampled steps of $V_{ij}$ share $\hat A_{ij}$, giving the following generator objective
\begin{equation}
\label{eq:generator_objective}
\begin{aligned}
\mathcal{J}_{\mathrm{G}}(\theta;\phi)
&=\mathbb{E}\Bigg[
\frac{1}{G l N}
\sum_{i=1}^{G}\sum_{j=1}^{l}\sum_{n=1}^{N}
\\[-2pt]
&\qquad\mathcal S_{\epsilon_{\mathrm G}}\!\left(
\rho_{ij,n}(\theta;\phi),\hat A_{ij}\right)
\Bigg].
\end{aligned}
\end{equation}
Both expectations are over the shared nested rollout. Following DanceGRPO~\citep{DBLP:journals/corr/abs-2505-07818}, $\mathcal J_{\mathrm G}$ has no KL term. At each optimization step, we compute both gradients before applying either update: $\nabla_\phi\mathcal J_{\mathrm R}$ through prompt likelihoods and $\nabla_\theta\mathcal J_{\mathrm G}$ through sampling transitions. The next rollout uses both updated policies, so the reasoner learns under the adapting generator, while the generator learns from evolving guidance and conditioning features. Algorithm~\ref{alg:codancerl} in the supplementary material summarizes the loop.

\subsection{Physics-Aware Feedback}
\label{sec:rewards}
We evaluate each generated video $V_{ij}$ against the real continuation $V^\star$ associated with its input condition. The physical-fidelity reward $r_{ij}$ measures agreement with this reference in appearance and motion regions, while the geometric trajectory reward $r'_{ij}$ measures agreement in point interactions. Both rewards serve as reference-based proxies for physical consistency. A different plausible continuation can therefore receive a lower score when its dynamics depart from the specific motion depicted in the reference video.

\paragraph{Physical-fidelity reward.}
We adapt the motion-overlap and reconstruction metrics of Physics-IQ Verified~\citep{DBLP:journals/corr/abs-2606-18943} to compare each generated video with a single real reference. The spatial (SP), spatiotemporal (ST), and weighted spatial (WS) IoU terms compare where motion occurs, whether it overlaps at the same time, and how frequently regions are active, respectively. MSE measures reconstruction error between videos with RGB values scaled to $[0,1]$. For each metric $M$, let $m_{ij}^{M}=M(V_{ij},V^\star)$. With larger IoU and smaller MSE indicating better agreement, we compute the raw physical-fidelity reward as the mean of all four terms
\begin{equation}
\label{eq:physical_reward}
r_{ij}=\frac{1}{4}\left[
1-m_{ij}^{\mathrm{MSE}}
+\sum_{M\in\{\mathrm{SP},\mathrm{ST},\mathrm{WS}\}}
m_{ij}^{M}
\right].
\end{equation}
The four terms lie in $[0,1]$ and are averaged with equal weights. We compare videos at a common spatial resolution and corresponding uniformly sampled times, excluding the shared conditioning frame from these reward computations.

\paragraph{Geometric trajectory reward.}
Similar motion regions can conceal different interactions between points. We therefore use CoTracker3~\citep{DBLP:journals/corr/abs-2410-11831} to track a common grid from the conditioning image in $V_{ij}$ and $V^\star$. Let $d,d^\star$ denote corresponding image-plane pairwise distances and $s$ the grid spacing, all normalized by the frame diagonal. The geometry score combines distance agreement with a soft-contact term:
\begin{equation}
\label{eq:trajectory_graph_reward}
m_{ij}^{\mathrm{geom}}
=1-\frac{1}{2}\mathbb{E}\!\left[
|d-d^\star|+\left|e^{-d/s}-e^{-d^\star/s}\right|
\right].
\end{equation}
Here, $\mathbb E$ averages jointly visible observations of pairs containing at least one moving point. The soft-contact term is more sensitive to distance differences when points are close. We also compare the times of closest approach through $m_{ij}^{\mathrm{time}}$, and final moving-point displacements after removing global translation through $m_{ij}^{\mathrm{final}}$. The final trajectory reward averages these three scores with equal weights:
\begin{equation}
\label{eq:trajectory_reward}
r'_{ij}=\frac{
m_{ij}^{\mathrm{geom}}+m_{ij}^{\mathrm{time}}+m_{ij}^{\mathrm{final}}
}{3}.
\end{equation}
Supplementary Section~\ref{app:reward-details} specifies point selection, visibility handling, and the timing and displacement scores.

\subsection{Hierarchical Reward Normalization}
\label{sec:normalization}
The nested rollout separates the choice of a rewrite from video generation under that rewrite. We assign credit at both levels. The generator compares videos within a fixed rewrite, while the reasoner compares rewrites through their mean raw video rewards. Prompt-level feedback uses these raw means because centering rewards within each rewrite makes their normalized mean zero and erases differences in mean reward between rewrites sampled for the same input condition.

At either level, the two reward families can differ in variance and favor different samples despite sharing the range $[0,1]$. We standardize each reward family separately before combining them with equal weights~\citep{DBLP:journals/corr/abs-2601-05242}. A final batch normalization controls the scale of the combined advantage. Batch statistics pool all input conditions in the same rollout batch, separately for videos and prompts.

\paragraph{Video-level advantage.}
For the $l$ videos under rewrite $p_i$, let $(\mu_i,\sigma_i)$ and $(\mu'_i,\sigma'_i)$ be the means and standard deviations of $\{r_{ij}\}_{j=1}^{l}$ and $\{r'_{ij}\}_{j=1}^{l}$. With stabilizer $\epsilon_{\mathrm{norm}}>0$, we compute the two standardized reward scores
\begin{equation}
\label{eq:video_group_normalization}
R_{ij}=\frac{r_{ij}-\mu_i}{\sigma_i+\epsilon_{\mathrm{norm}}},
\qquad
R'_{ij}=\frac{r'_{ij}-\mu'_i}{\sigma'_i+\epsilon_{\mathrm{norm}}}.
\end{equation}
Combining the standardized rewards and normalizing over the video batch gives the final video-level advantage
\begin{equation}
\label{eq:video_advantage}
\widetilde R_{ij}=\tfrac{1}{2}(R_{ij}+R'_{ij}),
\qquad
\hat{A}_{ij}=\frac{\widetilde R_{ij}-\mu_v}{\sigma_v+\epsilon_{\mathrm{norm}}},
\end{equation}
where $\mu_v$ and $\sigma_v$ are the batch mean and standard deviation of $\widetilde R_{ij}$. Thus, $\hat A_{ij}$ rewards a trajectory for improving on other videos generated from the same rewrite.

\paragraph{Prompt-level advantage.}
For each rewrite, the raw means $\bar r_i=\frac{1}{l}\sum_{j=1}^{l}r_{ij}$ and $\bar r'_i=\frac{1}{l}\sum_{j=1}^{l}r'_{ij}$ estimate its reward under the rollout generator. Let $(\mu_c,\sigma_c)$ and $(\mu'_c,\sigma'_c)$ be their means and standard deviations across the $G$ rewrites of the same input $c$. We compute the standardized scores
\begin{equation}
\label{eq:prompt_group_normalization}
R_i=\frac{\bar r_i-\mu_c}{\sigma_c+\epsilon_{\mathrm{norm}}},
\qquad
R'_i=\frac{\bar r'_i-\mu'_c}{\sigma'_c+\epsilon_{\mathrm{norm}}}.
\end{equation}
We then combine the standardized rewards and normalize over the full batch of sampled prompt rewrites:
\begin{equation}
\label{eq:prompt_advantage}
\widetilde R_i=\tfrac{1}{2}(R_i+R'_i),
\qquad
\hat{A}_i=\frac{\widetilde R_i-\mu_p}{\sigma_p+\epsilon_{\mathrm{norm}}},
\end{equation}
where $\mu_p$ and $\sigma_p$ are the batch mean and standard deviation of $\widetilde R_i$. The reasoner therefore favors rewrites whose videos achieve higher average physical rewards relative to other rewrites of $c$. Section~\ref{sec:ablation-study} evaluates normalization at both levels, with further analysis in Supplementary Section~\ref{app:normalization-analysis}.

\section{Experiments}
\label{sec:experiments}
We evaluate physical consistency on Physics-IQ Verified~\citep{DBLP:journals/corr/abs-2606-18943}, embodied video generation on RBench~\citep{DBLP:journals/corr/abs-2601-15282}, and the utility of learned features for closed-loop control on WorldArena~\citep{DBLP:journals/aiopen/ShangLMSJWJZTSGWLSZCZTC26}. Ablations examine coupled training, conditioning features, and hierarchical reward normalization.

\paragraph{Experimental setup.}
CoDance uses SFT followed by coupled RL on physical and embodied interaction videos, all disjoint from the evaluation data. Cosmos3-Nano, its SFT variant, and CoDance use identical video-generation settings within each benchmark. Data, training, evaluation protocols, and baseline result sources are detailed in Supplementary Sections~\ref{app:training-details} and~\ref{sec:experimental-setup}.

\subsection{Physically Plausible Image to Video Generation}
\label{sec:physically-plausible-image-to-video-generation}

\begin{table*}[t]
    \centering
    \begin{minipage}[t]{0.57\linewidth}
    \vspace{0pt}
    \caption{Comparison on Physics-IQ Verified~\citep{DBLP:journals/corr/abs-2606-18943} and a held-out VLM judge. MSE is reported as a reverse-normalized score, so higher is better. Bold and underlined values mark the best and second-best results, respectively.}
    \label{tab:physics-iq-verified}
    \footnotesize
    \setlength{\tabcolsep}{1.8pt}
    \renewcommand{\arraystretch}{1.12}
    \begin{tabularx}{\linewidth}{@{}>{\raggedright\arraybackslash}X*{6}{c}@{}}
        \toprule
        \multirow{2}{*}{\textbf{Model}} & \multicolumn{5}{c}{\textbf{Physics-IQ Verified}} & \multicolumn{1}{c}{\textbf{Held-out}} \\
        \cmidrule(lr){2-6}\cmidrule(l){7-7}
        & \textbf{S-IoU} & \textbf{ST-IoU} & \textbf{W-S-IoU} & \textbf{MSE} & \shortstack{\textbf{IQ-}\\\textbf{Score}} & \shortstack{\textbf{VLM-}\\\textbf{Judge}} \\
        \midrule
        \rowcolor{categoryblue}
        \multicolumn{7}{@{}l}{\textbf{\textit{Commercial}}} \\
        Sora 2~\citep{openai2025sora2}             & 37.3 & 27.0 & 26.9 & 14.8 & 26.5 & -- \\
        Grok-Imagine-Video~\citep{xai2026grokimagine} & 52.7 & 21.4 & 35.7 & 29.6 & 34.8 & -- \\
        MiniMax-H3 Max~\citep{fal2026h3max}     & 54.9 & 38.2 & 24.5 & 27.2 & 36.2 & - \\
        Seedance 2.0~\citep{DBLP:journals/corr/abs-2604-14148}       & 57.3 & 33.0 & \underline{41.6} & 24.6 & 39.1 & 41.9 \\
        \midrule
        \rowcolor{categoryblue}
        \multicolumn{7}{@{}l}{\textbf{\textit{Open-source}}} \\
        Wan2.2-I2V-A14B~\citep{DBLP:journals/corr/abs-2503-20314}    & 51.1 & 20.5 & 28.5 & 28.9 & 32.2 & 33.2 \\
        HunyuanVideo 1.5~\citep{DBLP:journals/corr/abs-2511-18870}   & 47.1 & 26.9 & 29.7 & 30.0 & 33.4 & 26.7 \\
        Cosmos3-Super I2V~\citep{DBLP:journals/corr/abs-2606-02800}  & 46.4   & \underline{49.3}   & 34.7   & 27.8   & 39.5 & 36.4 \\
        \midrule
        \rowcolor{categoryblue}
        \multicolumn{7}{@{}l}{\textbf{\textit{Physics-aware methods}}} \\
        Magi-1 24B+GeoPhys~\citep{DBLP:journals/corr/abs-2606-20707}
                                & --   & --   & --   & --   & 33.7 & -- \\
        PhiZero~\citep{DBLP:journals/corr/abs-2607-28624}            & \underline{57.4} & 36.3 & 26.8 & \textbf{41.7}   & \underline{40.6} & 33.9 \\
        \midrule
        \rowcolor{categoryblue}
        \multicolumn{7}{@{}l}{\textbf{\textit{Ours and base model}}} \\
        Cosmos3-Nano~\citep{DBLP:journals/corr/abs-2606-02800}       & 45.2 & 26.5 & 31.3 & \underline{36.3} & 34.8 & 31.7 \\
        Cosmos3-Nano + SFT & 44.3   & 38.0   & 32.5   & 35.6   & 37.6   & \underline{42.3} \\
        \rowcolor{ourspurple}
        \textbf{CoDance}      & \textbf{60.2} & \textbf{50.0} & \textbf{48.3} & 34.5 & \textbf{48.3} & \textbf{61.4} \\
        \bottomrule
    \end{tabularx}
    \end{minipage}\hfill
    \begin{minipage}[t]{0.41\linewidth}
        \vspace{0pt}
        \centering
        \includegraphics[width=\linewidth]{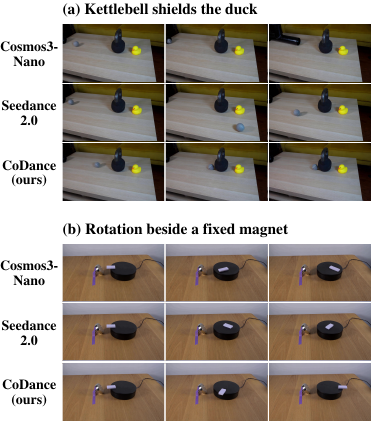}
        \captionof{figure}{\textbf{Qualitative comparison on Physics-IQ Verified.} Three temporally ordered frames per model show ball--kettlebell interaction and object rotation.}
        \label{fig:qualitative-comparison}
    \end{minipage}
\end{table*}

\begin{table*}[t]
    \centering
    \caption{Comparison of embodied video generation on RBench~\citep{DBLP:journals/corr/abs-2601-15282} across task types and embodiments. Avg. denotes the unweighted mean across the five task and four embodiment categories. The best and second-best results in each column are shown in bold and underlined, respectively. Task and embodiment categories correspond to separate evaluation subsets in RBench.}
    \label{tab:embodied-video-generation}
    \footnotesize
    \setlength{\tabcolsep}{2pt}
    \begin{tabularx}{\linewidth}{@{}>{\raggedright\arraybackslash}X*{10}{c}@{}}
        \toprule
        \multirow{2}{*}{\textbf{Model}} &
        \multicolumn{5}{c}{\textbf{Tasks}} &
        \multicolumn{4}{c}{\textbf{Embodiments}} &
        \multirow{2}{*}{\textbf{Avg.}} \\
        \cmidrule(lr){2-6}\cmidrule(lr){7-10}
        & \textbf{Manip.} & \textbf{Spatial} & \shortstack{\textbf{Multi-}\\\textbf{entity}} &
        \shortstack{\textbf{Long-}\\\textbf{horizon}} & \textbf{Reasoning} & \shortstack{\textbf{Single-}\\\textbf{arm}} &
        \shortstack{\textbf{Dual-}\\\textbf{arm}} & \textbf{Quadruped} & \textbf{Humanoid} & \\
        \midrule
        \rowcolor{categoryblue}
        \multicolumn{11}{@{}l}{\textbf{\textit{Open-source}}} \\
        Wan2.2-TI2V-5B~\citep{DBLP:journals/corr/abs-2503-20314} & 33.1 & 31.3 & 14.2 & 31.8 & 23.4 & 43.6 & 44.8 & 59.0 & 60.7 & 38.0 \\
        LTX-2~\citep{DBLP:journals/corr/abs-2601-03233} & 28.4 & 30.4 & 23.3 & 38.6 & 16.4 & 45.3 & 42.4 & 62.2 & 55.5 & 38.1 \\
        LongCat-Video~\citep{DBLP:journals/corr/abs-2510-22200} & 37.2 & 31.0 & 22.0 & 38.4 & 18.6 & 58.6 & 57.6 & 68.1 & 62.1 & 43.7 \\
        HunyuanVideo 1.5~\citep{DBLP:journals/corr/abs-2511-18870} & 44.2 & 31.6 & 31.2 & 43.8 & 36.4 & 51.3 & 52.6 & 63.4 & 59.5 & 46.0 \\
        Wan2.2-I2V-A14B~\citep{DBLP:journals/corr/abs-2503-20314} & 38.1 & 45.4 & 37.3 & 50.1 & 33.0 & 60.8 & 58.2 & 69.0 & 64.8 & 50.7 \\
        \midrule
        \rowcolor{categoryblue}
        \multicolumn{11}{@{}l}{\textbf{\textit{Commercial}}} \\
        Sora 2~\citep{openai2025sora2} & 20.8 & 26.8 & 18.6 & 25.5 & 11.5 & 47.6 & 51.3 & 66.4 & 56.1 & 36.1 \\
        Kling 2.6~\citep{kuaishou2025kling26} & 52.9 & 59.8 & 36.4 & 53.0 & 35.8 & 57.0 & 60.5 & 63.7 & 61.3 & 53.4 \\
        Hailuo v2~\citep{minimax2025hailuo02} & 56.0 & 63.7 & 38.6 & 54.5 & 47.4 & 59.4 & 61.1 & 64.0 & 63.5 & 56.5 \\
        Seedance 1.5~\citep{DBLP:journals/corr/abs-2512-13507} & 57.7 & 49.5 & 48.4 & 57.0 & 47.0 & 64.8 & 64.1 & 68.0 & 69.2 & 58.4 \\
        Seedance 2.0~\citep{DBLP:journals/corr/abs-2604-14148} & \underline{63.4} & 59.0 & \underline{54.7} & \underline{66.5} & \underline{54.5} & \textbf{72.8} & \textbf{73.4} & \textbf{78.2} & \textbf{73.8} & \underline{66.3} \\
        Veo 3.1~\citep{google2025veo31} & 54.1 & 47.4 & 53.4 & 59.2 & 46.7 & 67.0 & 66.6 & 74.3 & 70.4 & 59.9 \\
        Wan2.6~\citep{alibaba2025wan26} & 54.6 & 65.6 & 47.9 & 51.4 & 53.1 & 66.6 & 68.1 & 72.3 & 66.7 & 60.7 \\
        \midrule
        \rowcolor{categoryblue}
        \multicolumn{11}{@{}l}{\textbf{\textit{Robotics-specific}}} \\
        UnifoLM-WMA-0~\citep{unifolm-wma-0} & 3.6 & 4.0 & 1.8 & 6.2 & 0.0 & 26.8 & 19.4 & 29.3 & 20.0 & 12.3 \\
        Vidar~\citep{DBLP:journals/corr/abs-2507-12898} & 7.3 & 10.6 & 5.0 & 5.4 & 5.0 & 38.2 & 41.0 & 37.4 & 35.7 & 20.6 \\
        DreamGen (gr1)~\citep{DBLP:journals/corr/abs-2505-12705} & 31.2 & 37.2 & 29.7 & 33.4 & 21.5 & 56.4 & 53.2 & 57.9 & 57.5 & 42.0 \\
        Abot-PhysWorld~\citep{DBLP:journals/corr/abs-2603-23376} & 48.6 & 54.8 & 43.4 & 52.3 & 45.4 & 66.2 & 66.8 & 53.1 & 45.5 & 52.9 \\
        Cosmos3-Super~\citep{DBLP:journals/corr/abs-2606-02800} & 48.7 & 64.2 & 44.4 & 59.1 & 39.5 & 61.5 & 62.3 & 73.9 & 69.1 & 58.1 \\
        PF-Cosmos~\citep{DBLP:journals/corr/abs-2606-28128} & 58.9 & \underline{69.7} & 51.3 & 60.8 & 53.0 & 69.3 & 70.0 & 72.2 & 69.0 & 63.8 \\
        \midrule
        \rowcolor{categoryblue}
        \multicolumn{11}{@{}l}{\textbf{\textit{Ours and base model}}} \\
        Cosmos3-Nano~\citep{DBLP:journals/corr/abs-2606-02800} & 48.5 & 57.1 & 45.0 & 57.9 & 45.3 & 64.5 & 65.8 & 75.0 & 69.2 & 58.7 \\
        Cosmos3-Nano + SFT & 62.7 & 53.6 & 43.0 & 56.6 & 47.5 & \underline{71.8} & 71.5 & 74.2 & 72.2 & 61.5 \\
        \rowcolor{ourspurple}
        \textbf{CoDance} & \textbf{67.2} & \textbf{72.6} & \textbf{56.2} & \textbf{67.7} & \textbf{56.7} & 70.0 & \underline{71.8} & \underline{76.6} & \underline{72.3} & \textbf{67.9} \\
        \bottomrule
    \end{tabularx}

    \setlength{\intextsep}{8pt}
    \begin{minipage}{\linewidth}
        \centering
        \setlength{\abovecaptionskip}{4pt}
        \includegraphics[trim=9bp 7.4bp 5.25bp 5.25bp,clip,width=\linewidth]{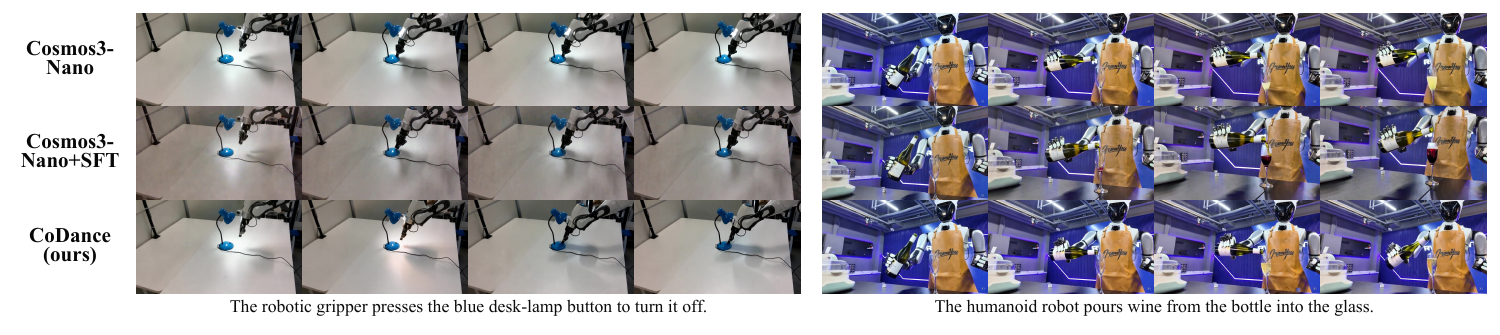}
        \captionof{figure}{\textbf{Qualitative comparison on RBench.} Rows compare Cosmos3-Nano, Cosmos3-Nano + SFT, and CoDance on two robotic interaction tasks, with columns within each case showing temporally ordered frames.}
        \label{fig:rbench-qualitative-comparison}
    \end{minipage}
\end{table*}

We first assess agreement with real physical continuations on Physics-IQ Verified (Table~\ref{tab:physics-iq-verified}). CoDance achieves the highest IQ-Score of 48.3, exceeding the base model by 13.5 points, SFT by 10.7 points, and PhiZero, the strongest external baseline, by 7.7 points. It leads all three IoU metrics, indicating better spatial and temporal overlap of motion with the reference. Reconstruction remains a weakness: its reverse-normalized MSE score decreases from 35.6 after SFT to 34.5, while PhiZero leads this metric.

Because the physical-fidelity reward incorporates these benchmark metrics, we also evaluate with a held-out VLM judge excluded from reward optimization. Its 0--100 score aggregates reference event fidelity, trajectory fidelity, physical plausibility, temporal coherence, and scene consistency. CoDance scores 61.4, compared with 31.7 for the base model, 42.3 for SFT, and 41.9 for Seedance 2.0, providing additional evidence of improved video quality beyond the optimized metrics. The judging protocol and full metric breakdown appear in Supplementary Sections~\ref{sec:experimental-setup} and~\ref{app:vlm-judge-breakdown}, respectively.

Figure~\ref{fig:qualitative-comparison} illustrates contact and motion constraints. In the displayed kettlebell sequence, CoDance shows the ball reaching and stopping against the obstacle, leaving the shielded duck undisturbed; the base model does not depict this contact. In the rotating-platform example, CoDance depicts the object moving with the platform while the adjacent magnet remains stationary, preserving the distinction between moving objects and fixed scene elements.

\subsection{Embodied Video Generation and Planning}
\label{sec:embodied-video-generation}

We next evaluate robotic interaction videos on RBench, covering five task categories and four embodiments (Table~\ref{tab:embodied-video-generation}). CoDance averages 67.9, compared with 58.7 for the base model, 61.5 for SFT, and 66.3 for Seedance 2.0. It leads all five task categories, while Seedance 2.0 leads all four embodiment categories. Relative to SFT, all task scores improve, with an unweighted mean gain of 11.40 points. The mean gain across embodiment categories is only 0.25 points, and single-arm performance declines. This pattern suggests that post-training primarily improves task-conditioned interactions, with limited gains across embodiment categories.

Figure~\ref{fig:rbench-qualitative-comparison} illustrates two task-relevant state transitions. The base and SFT sequences leave the lamp lit, while CoDance pairs the button press with the light turning off. In the pouring example, CoDance depicts the bottle returning toward an upright pose after tilting over the glass; the baseline sequences end with the bottle still tilted. These examples connect the depicted actions to their effects and progression toward completion of the intended interaction tasks.

\begin{table*}[t]
    \centering
    \begin{minipage}[t]{0.44\linewidth}
        \vspace{0pt}
        \centering
        \setlength{\abovecaptionskip}{0pt}
        \setlength{\belowcaptionskip}{4pt}
        \caption{WorldArena closed-loop action-planning success rates (\%) for each task and their mean.}
        \label{tab:worldarena-action-planning}
        \footnotesize
        \setlength{\tabcolsep}{1.5pt}
        \begin{tabularx}{\linewidth}{@{}>{\raggedright\arraybackslash}X*{3}{>{\raggedleft\arraybackslash}p{0.16\linewidth}}@{}}
            \toprule
            \textbf{Model} & \shortstack[r]{\textbf{Adjust}\\\textbf{bottle}} & \shortstack[r]{\textbf{Click}\\\textbf{bell}} & \textbf{Avg.} \\
            \midrule
            RoboMaster~\citep{DBLP:conf/iclr/FuWLBXWZL26}            &  8.0 & 20.0 & 14.0 \\
            Wan2.2-TI2V-5B~\citep{DBLP:journals/corr/abs-2503-20314}        & 12.0 & 20.0 & 16.0 \\
            TesserAct~\citep{DBLP:journals/corr/abs-2504-20995}             &  1.0 & \underline{35.0} & 18.0 \\
            WoW~\citep{DBLP:journals/corr/abs-2509-22642}                   & 20.0 & 21.0 & 20.5 \\
            PhysisForcing~\citep{DBLP:journals/corr/abs-2606-28128}         & \underline{22.0} & 26.0 & \underline{24.0} \\
            \midrule
            Cosmos3-Nano~\citep{DBLP:journals/corr/abs-2606-02800}          & 15.0 & 13.0 & 14.0 \\
            Cosmos3-Nano+SFT              &  21.0 & 24.0 & 22.5 \\
            \rowcolor{ourspurple}
            \textbf{CoDance}    & \textbf{34.0} & \textbf{49.0} & \textbf{41.5} \\
            \bottomrule
        \end{tabularx}
    \end{minipage}%
    \hfill%
    \begin{minipage}[t]{0.54\linewidth}
        \vspace{0pt}
        \setlength{\intextsep}{0pt}
        \begin{minipage}{\linewidth}
            \centering
            \setlength{\abovecaptionskip}{8pt}
            \includegraphics[trim=26.5bp 10.25bp 0bp 0bp,clip,width=\linewidth]{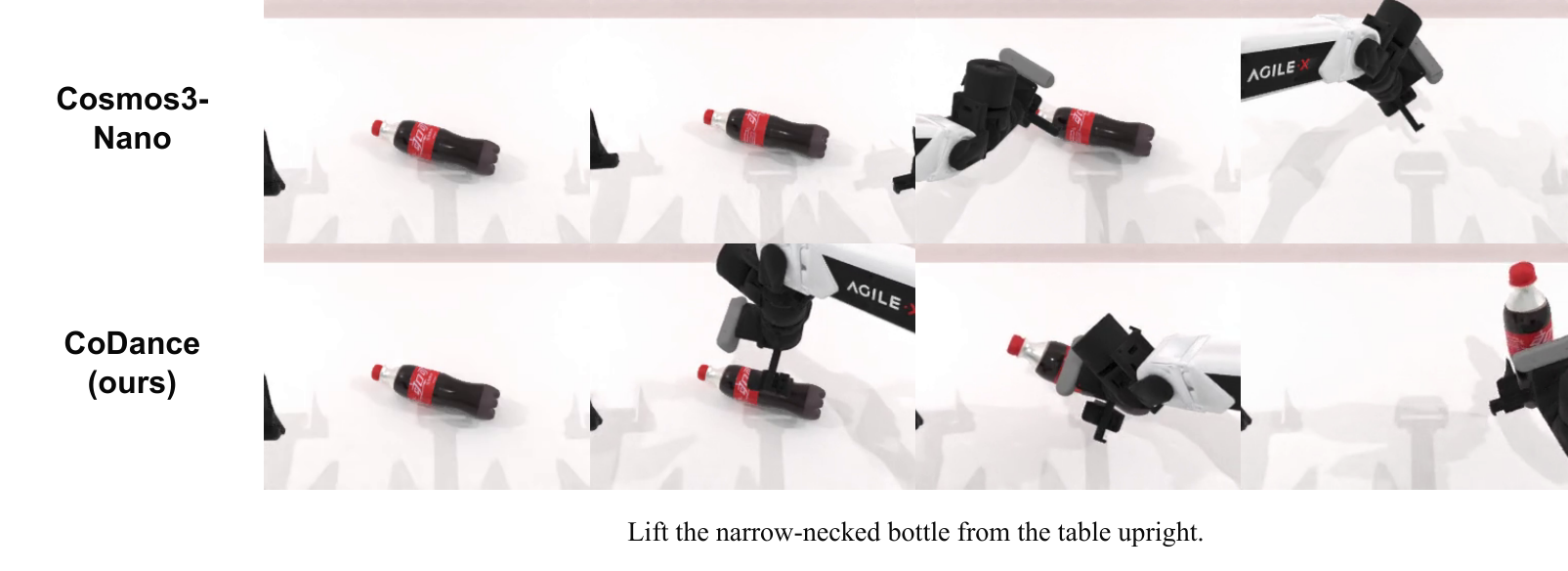}
            \captionof{figure}{\textbf{Qualitative comparison on adjust\_bottle.} The controller using CoDance features grasps the bottle and holds it upright; the controller using Cosmos3-Nano features does not reach this pose.}
            \label{fig:bottle-lifting-comparison}
        \end{minipage}
    \end{minipage}
\end{table*}

We evaluate closed-loop control on WorldArena's \texttt{adjust\_bottle} and \texttt{click\_bell} tasks. With CoDance frozen, the reasoner generates a rewrite that conditions the generator. A task-specific inverse dynamics model (IDM) maps the generator's features to action chunks, without synthesizing future videos. CoDance's SFT and RL data exclude RoboTwin 2.0. Across 100 rollouts per task (ten initial conditions, ten repetitions each), CoDance achieves success rates of 34.0\% and 49.0\% (Table~\ref{tab:worldarena-action-planning}). Mean success reaches 41.5\%, compared with 14.0\% for the base model and 22.5\% for SFT, supporting the transfer of reasoner-guided generator features to control. See Figure~\ref{fig:bottle-lifting-comparison} for bottle lifting and Supplementary Table~\ref{tab:additional-robotwin-tasks} for results on three more RoboTwin 2.0 tasks.

\subsection{Ablation and Analysis}
\label{sec:ablation-study}

Table~\ref{tab:ablation-overview} tests whether the gains depend on coupled updates, adapted conditioning features, and normalization at both rollout levels on Physics-IQ Verified.

\begin{table*}[t]
    \centering
    \caption{Ablations on Physics-IQ Verified, measured by IQ-Score ($\uparrow$). (a) Training strategies evaluated with a shared base encoder or each model's own conditioning encoder, holding rewrites and the generator fixed within each row. $\Delta$ is Model minus Base. (b) Reward-decoupled normalization at the video and prompt levels, with both physical rewards enabled in all four settings. Gain is relative to no normalization.}
    \label{tab:ablation-overview}
    \small
    \setlength{\tabcolsep}{2pt}
    \renewcommand{\arraystretch}{1.0}
    \begin{minipage}[t]{\columnwidth}
        \vspace{0pt}
        \centering
        \textbf{\strut (a) Training strategies and encoders}\par\smallskip
        \begin{tabularx}{\linewidth}{@{}>{\raggedright\arraybackslash}Xrrr@{}}
            \toprule
            \multirow{2}{*}{\textbf{Training strategy}} & \multicolumn{2}{c}{\textbf{Conditioning encoder}} & \multirow{2}{*}{$\boldsymbol{\Delta}$} \\
            \cmidrule(lr){2-3}
            & \textbf{Base} & \textbf{Model} & \\
            \midrule
            Supervised fine-tuning (SFT)     & 38.4 & 37.6 & $-0.8$ \\
            GRPO (reasoner-only)            & 41.5 & 42.0 & $+0.5$ \\
            DanceGRPO (generator-only)     & 39.3 & 39.3 & $0.0$ \\
            GRPO + DanceGRPO (sequential) & 44.1 & 44.2 & $+0.1$ \\
            \midrule
            \rowcolor{ourspurple}
            \textbf{CoDanceRL (joint training)} & \textbf{45.4} & \textbf{48.3} & $\boldsymbol{+2.9}$ \\
            \bottomrule
        \end{tabularx}
    \end{minipage}\hfill
    \begin{minipage}[t]{\columnwidth}
        \vspace{0pt}
        \centering
        \textbf{\strut (b) Normalization levels}\par\smallskip
        \newcommand{\normrowstrut}{\vrule height 1.333333\ht\strutbox depth 1.333333\dp\strutbox width 0pt}
        \begin{tabularx}{\linewidth}{@{}>{\centering\arraybackslash}X>{\centering\arraybackslash}Xrr@{}}
            \toprule
            \multicolumn{2}{c}{\textbf{Normalization level}} & \multirow{2}{*}{\textbf{IQ-Score}$\uparrow$} & \multirow{2}{*}{\textbf{Gain}} \\
            \cmidrule(lr){1-2}
            \textbf{Video} & \textbf{Prompt} & & \\
            \midrule
            \normrowstrut\xmark & \xmark & 41.3 & $0.0$ \\
            \normrowstrut\cmark & \xmark & 43.9 & $+2.6$ \\
            \normrowstrut\xmark & \cmark & 43.3 & $+2.0$ \\
            \midrule
            \rowcolor{ourspurple}
            \cmark & \cmark & \textbf{48.3} & $\boldsymbol{+7.0}$ \\
            \bottomrule
        \end{tabularx}
    \end{minipage}
\end{table*}

\paragraph{Comparison of Training Strategies.}
All RL variants share the SFT initialization, both physical rewards, and reward-decoupled normalization, with matched rollout and update counts. Using each model's own encoder, joint training reaches 48.3, compared with 42.0 for reasoner-only RL and 39.3 for generator-only RL. Sequential RL first trains the reasoner and then the generator, reaching 44.2. CoDanceRL's 4.1-point advantage supports continued adaptation of both policies throughout RL. Supplementary Section~\ref{app:training-component-ablation} provides the full breakdown of all evaluation metrics.

\paragraph{Effect of the Conditioning Encoder.}
Within each row of Table~\ref{tab:ablation-overview}(a), we hold the rewrite text and generator fixed and replace only the conditioning encoder. Using CoDance's encoder raises IQ-Score from 45.4 to 48.3, a gain of 2.9 points, compared with gains of 0.1 for sequential RL and 0.5 for reasoner-only RL. This gain supports improved compatibility between conditioning features and the generator under coupled training. Supplementary Section~\ref{app:feature-generator-compatibility} provides further component recombinations at inference time.

\paragraph{Ablation of Normalization Levels.}
With both physical rewards fixed, Table~\ref{tab:ablation-overview}(b) isolates the normalization levels. Applying reward-decoupled normalization at the video or prompt level raises IQ-Score from 41.3 to 43.9 or 43.3, respectively. Using both levels reaches 48.3, exceeding the two partial configurations by 4.4 and 5.0 points. These results support constructing normalized feedback at both levels of the rollout hierarchy. Supplementary Section~\ref{app:normalization-analysis} reports the complete reward and normalization ablations.

\section{Related Work}
\label{sec:related-work}
\textbf{Physically Consistent Video Generation.}
Existing methods improve physical consistency by grounding generation in simulation~\citep{DBLP:conf/eccv/LiuRGW24}, adapting generators with physical supervision~\citep{DBLP:journals/corr/abs-2603-13770,DBLP:journals/corr/abs-2606-28128,DBLP:journals/corr/abs-2604-02817}, or using physics-aware preferences and rewards~\citep{DBLP:conf/eccv/CaiLJWSLCJWTDJYH26,DBLP:journals/corr/abs-2512-00425}.
Reasoning-based approaches make interactions explicit through prompt refinement~\citep{DBLP:conf/cvpr/Xue00025,DBLP:journals/corr/abs-2511-20280} or structured keyframes and trajectories~\citep{DBLP:journals/corr/abs-2606-14317}, with generator weights fixed.
Generation feedback can also train the source of this guidance: PhyPrompt optimizes a rewriter with a frozen generator~\citep{DBLP:journals/corr/abs-2603-03505}, while PhysMaster uses joint supervised fine-tuning followed by separate preference-optimization stages for the physical encoder and generator~\citep{DBLP:journals/corr/abs-2510-13809}.
CoDance keeps the reasoner and generator trainable throughout RL, so the generator can adapt to evolving physical guidance and the conditioning features produced by the reasoner.

\textbf{Reinforcement Learning for Video Generation.}
DanceGRPO applies GRPO to visual generators~\citep{DBLP:journals/corr/abs-2505-07818}, and TAGRPO refines optimization through reward-ranked trajectory alignment~\citep{DBLP:journals/corr/abs-2601-05729}.
Feedback includes learned preference rewards~\citep{DBLP:journals/corr/abs-2509-08826}, motion-control objectives~\citep{DBLP:conf/iclr/ZhaoSZZSZSZ26}, and identity-preservation rewards~\citep{DBLP:conf/cvpr/ShenJZLGCZ26}.
VANS aligns a VLM and video generator through sequential training for next-event prediction~\citep{DBLP:conf/cvpr/ChengHTL26}.
CoDance focuses on physical I2V generation, with the reasoner also acting as the conditioning encoder. Shared physical-fidelity and geometric trajectory rewards supervise both policies. Rewrite-level advantages reflect the quality of multiple video outcomes, while video-level advantages distinguish individual video realizations generated under the same rewritten guidance.

\section{Conclusion}
\label{sec:conclusion}
We presented \textbf{CoDance} and \textbf{CoDanceRL} to improve how physical reasoning guides video generation. Coupled training uses shared physical feedback to learn both the guidance and its realization as motion and interactions. Improvements on Physics-IQ Verified and RBench, together with held-out VLM judgments, support the effectiveness of this approach. Controlled ablations further support adapting both components and normalizing rewards at both rollout levels. Beyond generation, frozen reasoner-guided generator features achieve 41.5\% average success across two WorldArena tasks, providing evidence of transfer to closed-loop control without synthesizing future videos.

Precise reference-motion matching remains challenging. Our single-reference rewards can penalize physically valid alternatives, while reference agreement does not guarantee physical correctness. Future work could develop feedback that accommodates multiple plausible futures and evaluate transfer across a broader range of control tasks.

\label{paper:maintext-end}
\clearpage
\section*{LLM Usage Statement}
Generative AI tools were used in research execution and to generate synthetic
training data in two specific roles. First, GPT-5.6 Sol generated physically
informed prompt rewrites that served as supervised training targets for
reasoner fine-tuning, as described in
Supplementary Section~\ref{app:dataset-construction}. Second, GPT-5.6 Luna and GPT-5
performed automated model evaluation as the VLM judges for Physics-IQ
Verified and RBench, respectively, as described in
Supplementary Section~\ref{sec:experimental-setup}. These judges were excluded from reward
optimization. OpenAI Codex was also used to improve the grammar, clarity,
and readability of author-written text and to help locate potentially relevant
literature and source material. They were not used for research ideation, to
originate scientific claims or technical content, to formulate or prove
mathematical claims, or for other research tasks requiring disclosure beyond
the roles stated above. All AI-assisted work was conducted under author
supervision and reviewed before inclusion. The authors verified candidate
references against their original sources and checked the AI-assisted
experimental outputs and their aggregation. The authors take responsibility for
the final content of this work, including its text, claims, citations, data, and
reported results.

%% file: sections/appendix.tex
\clearpage
\setcounter{page}{1}
\renewcommand{\thepage}{S\arabic{page}}
\maketitlesupplementary
\appendix
\section{Method Details}
\label{app:method-details}

\subsection{CoDanceRL Algorithm Pipeline}
\label{app:codancerl-pipeline}

Algorithm~\ref{alg:codancerl} summarizes CoDanceRL using the notation in Section~\ref{sec:method}. We omit the input-condition index for readability. The final batch normalization pools samples from all conditions in the rollout batch.

\begin{algorithm*}[p]
    \caption{CoDanceRL}
    \label{alg:codancerl}
    \small
    \algrenewcommand{\algorithmicindent}{1em}
    \algrenewcommand{\algorithmiccomment}[1]{\hfill\textit{// #1}}
    \begin{algorithmic}[1]
        \State \textbf{Input}
        \State \hspace{1em}1) Dataset $\mathcal D$ containing conditions $c=(I,p)$ and reference videos $V^\star$
        \State \hspace{1em}2) Reasoner $\pi_\phi$, conditioning encoder $E_\phi$, generator $p_\theta$, and frozen reference $\pi_{\phi_{\mathrm{ref}}}$
        \State \hspace{1em}3) Rewrites per input $G$, videos per rewrite $l$, and sampling steps $N$
        \State \hspace{1em}4) Clipping thresholds $\epsilon_{\mathrm R},\epsilon_{\mathrm G}$, dual clip $\kappa$, stabilizer $\epsilon_{\mathrm{norm}}>0$, KL weight $\beta$, learning rates $\eta_\phi,\eta_\theta$
        \State \textbf{Initialize} parameters $\phi,\theta$ from supervised fine-tuning
        \For{each training iteration}
            \State \textbf{Freeze old policies} by setting $\phi_{\mathrm{old}}\gets\phi$, $\theta_{\mathrm{old}}\gets\theta$
            \State Sample a batch $\mathcal C\subset\mathcal D$ and initialize rollout buffer $\mathcal B\gets\emptyset$
            \State \textbf{Nested rollout collection}
            \For{each $(c,V^\star)\in\mathcal C$}
                \State Sample rewrites $\{p_i\}_{i=1}^{G}\sim\pi_{\phi_{\mathrm{old}}}(\cdot\mid c)$
                \For{$i=1$ to $G$}
                    \State Set $c_i\gets(I,p_i)$ and conditioning features $h_i^{\mathrm{old}}\gets E_{\phi_{\mathrm{old}}}(c_i)$
                    \For{$j=1$ to $l$}
                        \State Sample $\xi_{ij}\sim\mathcal N(0,\mathbf I)$ and set $z_{ij,N}\gets\xi_{ij}$
                        \State Sample $z_{ij,n-1}\sim p_{\theta_{\mathrm{old}}}(\cdot\mid z_{ij,n},I,h_i^{\mathrm{old}})$ for $n=N,\ldots,1$
                        \State Decode $z_{ij,0}$ into video $V_{ij}$
                        \State Compute rewards $r_{ij},r'_{ij}$ using Eqs.~\ref{eq:physical_reward} and~\ref{eq:trajectory_reward}
                    \EndFor
                    \State \textbf{Video-level reward normalization}
                    \State Compute means/stds $(\mu_i,\sigma_i)$ of $\{r_{ij}\}_{j=1}^{l}$ and $(\mu'_i,\sigma'_i)$ of $\{r'_{ij}\}_{j=1}^{l}$
                    \State $R_{ij}\gets\frac{r_{ij}-\mu_i}{\sigma_i+\epsilon_{\mathrm{norm}}}$, $\quad R'_{ij}\gets\frac{r'_{ij}-\mu'_i}{\sigma'_i+\epsilon_{\mathrm{norm}}}$ for all $j$
                    \State $\widetilde R_{ij}\gets\tfrac{1}{2}(R_{ij}+R'_{ij})$ for all $j$
                    \State Compute raw reward means $\bar r_i\gets\frac{1}{l}\sum_{j=1}^{l}r_{ij}$, $\quad\bar r'_i\gets\frac{1}{l}\sum_{j=1}^{l}r'_{ij}$
                \EndFor
                \State \textbf{Prompt-level reward normalization}
                \State Compute means/stds $(\mu_c,\sigma_c)$ of $\{\bar r_i\}_{i=1}^{G}$ and $(\mu'_c,\sigma'_c)$ of $\{\bar r'_i\}_{i=1}^{G}$
                \State $R_i\gets\frac{\bar r_i-\mu_c}{\sigma_c+\epsilon_{\mathrm{norm}}}$, $\quad R'_i\gets\frac{\bar r'_i-\mu'_c}{\sigma'_c+\epsilon_{\mathrm{norm}}}$ for all $i$
                \State $\widetilde R_i\gets\tfrac{1}{2}(R_i+R'_i)$ for all $i$
                \State Store conditions, rewrites, latent trajectories, rollout log-probabilities, and $\widetilde R_i,\widetilde R_{ij}$ in $\mathcal B$
            \EndFor
            \State \textbf{Batch normalization}
            \State Compute batch means/stds $(\mu_v,\sigma_v)$ of $\widetilde R_{ij}$ and $(\mu_p,\sigma_p)$ of $\widetilde R_i$
            \State Assign advantages $\hat A_{ij}\gets\frac{\widetilde R_{ij}-\mu_v}{\sigma_v+\epsilon_{\mathrm{norm}}}$, $\quad\hat A_i\gets\frac{\widetilde R_i-\mu_p}{\sigma_p+\epsilon_{\mathrm{norm}}}$ to all stored rollouts
            \State \textbf{Coupled policy optimization}
            \For{each optimization step on $\mathcal B$}
                \State \textbf{Compute response-level reasoner ratio} $\rho_i(\phi)\gets\frac{\pi_\phi(p_i\mid c)}{\pi_{\phi_{\mathrm{old}}}(p_i\mid c)}$
                \State \textbf{Compute generator ratio} $\rho_{ij,n}(\theta;\phi)$ using Eq.~\ref{eq:coupled_generator_ratio} and the stored rollout log-probabilities
                \State \textbf{Compute both gradients before either update:} $g_\phi\gets\nabla_\phi\mathcal J_{\mathrm R}(\phi)$, $\quad g_\theta\gets\nabla_\theta\mathcal J_{\mathrm G}(\theta;\phi)$
                \State Hold $\phi$ fixed in $g_\theta$; treat stored rewards, advantages, and rollout log-probabilities as constants
                \State \textbf{Update both policies:} $\phi\gets\phi+\eta_\phi g_\phi$, $\quad\theta\gets\theta+\eta_\theta g_\theta$
            \EndFor
        \EndFor
        \State \textbf{Return} the trained reasoner $\pi_\phi$ and video generator $p_\theta$
    \end{algorithmic}
\end{algorithm*}

\subsection{Geometric Trajectory Reward Computation}
\label{app:reward-details}

CoTracker3 tracks a common grid of points from the conditioning image in $V_{ij}$ and $V^\star$. Spatial quantities, including grid spacing $s$, are normalized by the frame diagonal. A point is moving if its maximum reference displacement after removing median global translation exceeds $s$. We use only observations visible in both videos after the conditioning frame. Unstarred and starred quantities below refer to $V_{ij}$ and $V^\star$, respectively. Here, $\mathbb{E}$ is the uniform empirical average over the valid observations for each score.

For geometry, we consider pairs containing at least one moving point. Their normalized distances $d,d^\star\in[0,1]$ at jointly visible frames enter Eq.~\ref{eq:trajectory_graph_reward}. Its empirical average gives equal weight to valid pair--frame observations; $e^{-d/s}$ increases as points approach.

For timing, we retain these pairs when their reference distance changes by more than $s$. Let $\tau,\tau^\star\in[0,1]$ be their times of closest approach, normalized by video duration. Averaging over the retained pairs gives
\begin{equation}
\label{eq:trajectory_order_reward}
m_{ij}^{\mathrm{time}}
=1-\mathbb{E}\!\left[|\tau-\tau^\star|\right].
\end{equation}

For final displacement, let $u,u^\star$ be a moving point's initial-to-final displacements after subtracting each video's median point displacement. We cap the displacement error at one and average over moving points visible in both final frames to obtain
\begin{equation}
\label{eq:trajectory_terminal_reward}
m_{ij}^{\mathrm{final}}
=1-\mathbb{E}\!\left[
\min\!\left(\|u-u^\star\|_2,1\right)
\right].
\end{equation}
The three scores are averaged in Eq.~\ref{eq:trajectory_reward}.

\section{Dataset and Training Details}
\label{app:training-details}

\subsection{Dataset Construction Details}
\label{app:dataset-construction}

\paragraph{RL training data.}
Our RL training dataset combines videos of physical phenomena with embodied interaction videos. To support physically consistent video generation, we select approximately 1,000 videos from OpenVid-1M~\citep{DBLP:conf/iclr/NanXZFYCL0T25} depicting physical phenomena captured with a static camera, and supplement them with 254 real-world videos that we record to cover a range of physical phenomena. To support embodied video generation, we select approximately 3,000 high-quality videos from RoVid-X~\citep{DBLP:journals/corr/abs-2601-15282} featuring rich physical interactions. All training data are disjoint from the test data used in our evaluations.

\paragraph{SFT training data.}
The generator is fine-tuned separately on the full RL video dataset. Each training example uses the video's first frame $I$ and original prompt $p$ as the condition, and the corresponding ground-truth video $V^\star$ as the target. For the reasoner, we use GPT-5.6 Sol~\citep{openai2026gpt56sol} with the \texttt{max} reasoning setting to produce a physically informed rewrite of the original prompt for every video in the same dataset. The reasoner takes the first frame $I$ and original prompt $p$ as input and is trained with teacher forcing to predict the teacher-generated rewrite.

\subsection{Supervised Fine-Tuning}
\label{app:sft-details}
We fine-tune the reasoner and generator separately on the dataset in Section~\ref{app:dataset-construction}, then use both checkpoints to initialize coupled reinforcement learning.

\paragraph{Reasoner SFT.}
The reasoner receives the first frame $I$ and original prompt $p$ and learns to predict the teacher-generated physical rewrite. We minimize the token-level negative log-likelihood with teacher forcing, updating LoRA adapters in the attention and feed-forward projections while keeping the base weights and vision encoder frozen.

\paragraph{Generator SFT.}
We condition the generator on the first frame $I$ and original prompt $p$, using the corresponding reference video $V^\star$ as the target. The reference video is encoded into latent space for the flow-matching objective in Section~\ref{sec:preliminary}. Training updates the video-generation parameters while keeping the language-conditioning tower, vision encoder, and video autoencoder frozen.

\subsection{Reinforcement Learning}
\label{app:rl-details}
We report the parameter configuration for the full nested rollout protocol in Section~\ref{sec:optimization}, with $G=8$ rewrites per input and $l=8$ videos per rewrite.

\paragraph{Rollout configuration.}
With the default batch size of $16$ input conditions, each iteration contains $128$ prompt rewrites and $1{,}024$ generated videos. The reasoner samples rewrites with temperature $0.7$, top-$p=0.8$, top-$k=20$, a repetition penalty of $1.0$, and a presence penalty of $1.5$. The maximum completion length is $8{,}192$ tokens. Each of the eight rewrites conditions a separate group of eight videos. The physical-fidelity and geometric trajectory rewards are converted into video-level and prompt-level advantages using the hierarchical normalization in Section~\ref{sec:normalization}.

\paragraph{Policy optimization.}
The reasoner and generator use separate AdamW optimizers with constant learning rates of $3\times10^{-6}$ and $1\times10^{-5}$, and weight decay of $10^{-2}$ and $10^{-4}$, respectively. Both optimizers use an Adam numerical-stability constant of $10^{-8}$ and a maximum gradient norm of $1.0$. The PPO clipping thresholds are $\epsilon_{\mathrm R}=0.2$ and $\epsilon_{\mathrm G}=10^{-4}$. With dual-clipping coefficient $\kappa=3.0$, the reasoner's surrogate in Eq.~\ref{eq:reasoner_objective} is
\begin{equation}
\label{eq:reasoner_dual_clip}
\mathcal S_{\mathrm R}(\rho,A)=
\begin{cases}
\max\!\left(\mathcal S_{\epsilon_{\mathrm R}}(\rho,A),\kappa A\right), & A<0,\\
\mathcal S_{\epsilon_{\mathrm R}}(\rho,A), & A\geq0.
\end{cases}
\end{equation}
Its KL penalty is computed against the frozen SFT adapter using the $k_3$ estimator with coefficient $\beta=0.03$. The generator follows Eq.~\ref{eq:generator_objective}.

\subsection{Training Dynamics}
\label{app:training-dynamics}
\begin{figure*}[t]
    \centering
    \includegraphics[width=\linewidth]{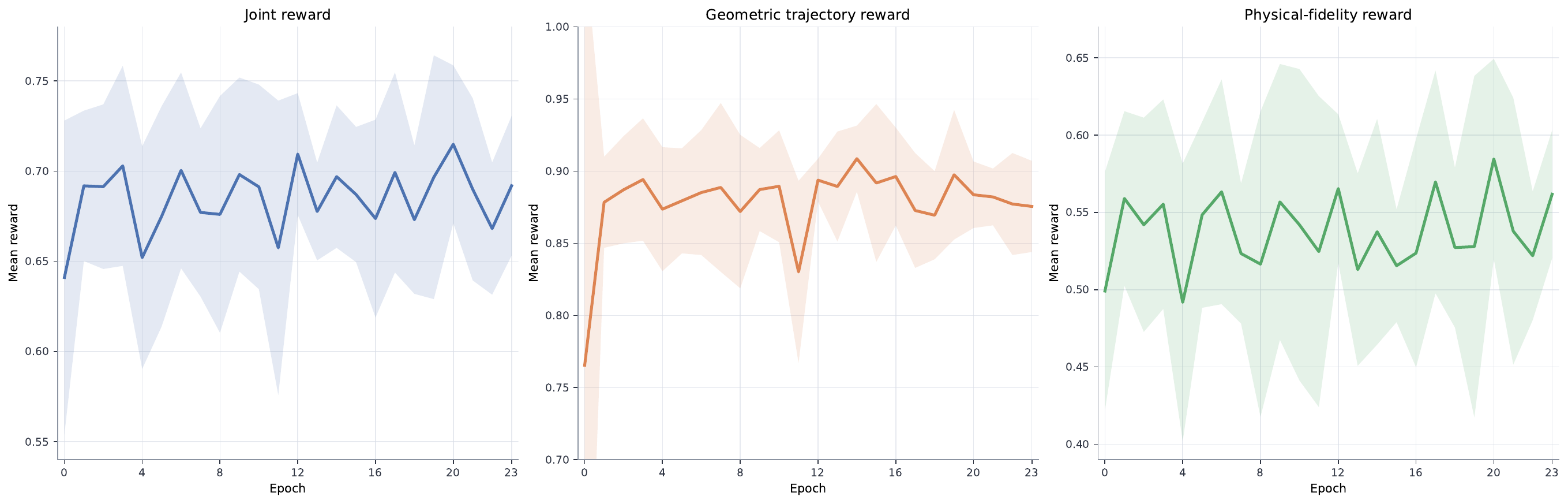}
    \caption{\textbf{Training dynamics of CoDanceRL.} Mean joint reward (left), geometric trajectory reward (middle), and physical-fidelity reward (right) over training epochs 0--23. Shaded bands indicate standard deviations.}
    \label{fig:training-dynamics}
\end{figure*}

Figure~\ref{fig:training-dynamics} presents the mean joint reward and its two components over epochs 0--23. The joint and geometric trajectory rewards rise mainly during the initial epochs and then fluctuate around higher levels. The physical-fidelity reward fluctuates throughout training but ends above its initial value.

\section{Evaluation Details}
\label{sec:experimental-setup}

\paragraph{Compared models.}
Tables~\ref{tab:physics-iq-verified}, \ref{tab:embodied-video-generation}, and~\ref{tab:worldarena-action-planning} compare CoDance with general-purpose video models, physics-aware methods, and embodied world models. The general-purpose video baselines include Sora 2~\citep{openai2025sora2}, Grok-Imagine-Video~\citep{xai2026grokimagine}, Seedance 1.5~\citep{DBLP:journals/corr/abs-2512-13507} and 2.0~\citep{DBLP:journals/corr/abs-2604-14148}, Kling 2.6~\citep{kuaishou2025kling26}, Hailuo v2~\citep{minimax2025hailuo02}, Veo 3.1~\citep{google2025veo31}, Wan2.6~\citep{alibaba2025wan26}, Wan2.2-I2V-A14B~\citep{DBLP:journals/corr/abs-2503-20314}, Wan2.2-TI2V-5B~\citep{DBLP:journals/corr/abs-2503-20314}, HunyuanVideo 1.5~\citep{DBLP:journals/corr/abs-2511-18870}, MiniMax-H3 Max~\citep{fal2026h3max}, LTX-2~\citep{DBLP:journals/corr/abs-2601-03233}, and LongCat-Video~\citep{DBLP:journals/corr/abs-2510-22200}. The physics-aware and robotics-oriented baselines include Magi-1 24B+GeoPhys~\citep{DBLP:journals/corr/abs-2505-13211,DBLP:journals/corr/abs-2606-20707}, PhiZero~\citep{DBLP:journals/corr/abs-2607-28624}, UnifoLM-WMA-0~\citep{unifolm-wma-0}, Vidar~\citep{DBLP:journals/corr/abs-2507-12898}, DreamGen (gr1)~\citep{DBLP:journals/corr/abs-2505-12705}, Abot-PhysWorld~\citep{DBLP:journals/corr/abs-2603-23376}, Cosmos3-Super I2V~\citep{DBLP:journals/corr/abs-2606-02800}, Cosmos3-Super~\citep{DBLP:journals/corr/abs-2606-02800}, and PF-Cosmos~\citep{DBLP:journals/corr/abs-2606-28128}. For closed-loop planning, we compare with RoboMaster~\citep{DBLP:conf/iclr/FuWLBXWZL26}, Wan2.2-TI2V-5B~\citep{DBLP:journals/corr/abs-2503-20314}, TesserAct~\citep{DBLP:journals/corr/abs-2504-20995}, WoW~\citep{DBLP:journals/corr/abs-2509-22642}, and PhysisForcing~\citep{DBLP:journals/corr/abs-2606-28128}. We also include Cosmos3-Nano~\citep{DBLP:journals/corr/abs-2606-02800} and our Cosmos3-Nano + SFT variant to measure the gains from supervised initialization and subsequent reinforcement learning. In Table~\ref{tab:ablation-study}, GRPO~\citep{DBLP:journals/corr/abs-2402-03300} and DanceGRPO~\citep{DBLP:journals/corr/abs-2505-07818} provide reasoner-only and generator-only training baselines, respectively, while component recombinations in Table~\ref{tab:component-composition} evaluate the contributions of the trained reasoner and generator.

\paragraph{Result sources.}
For Physics-IQ Verified (Table~\ref{tab:physics-iq-verified}), we evaluate all models in the \textit{Open-source} and \textit{Ours and base model} groups, as well as MiniMax-H3 Max, Seedance 2.0, and PhiZero. The remaining results, for Sora 2, Grok-Imagine-Video, and Magi-1 24B+GeoPhys, are quoted from the Physics-IQ Verified paper~\citep{DBLP:journals/corr/abs-2606-18943} and released GitHub results.\footnote{\url{https://github.com/google-deepmind/physics-IQ-benchmark}.}

For RBench (Table~\ref{tab:embodied-video-generation}), we evaluate Seedance 1.5, Seedance 2.0, Cosmos3-Nano, Cosmos3-Nano + SFT, and CoDance. Category scores for all entries in the \textit{Open-source} and \textit{Robotics-specific} groups, and for the remaining \textit{Commercial} entries, are quoted from Table~1 of PhysisForcing~\citep{DBLP:journals/corr/abs-2606-28128}.

For WorldArena (Table~\ref{tab:worldarena-action-planning}), we evaluate Cosmos3-Nano, Cosmos3-Nano + SFT, and CoDance. Results for the other planners are quoted from Table~3 of PhysisForcing~\citep{DBLP:journals/corr/abs-2606-28128}.

\paragraph{VLM-as-judge evaluations.}
We use GPT-5.6 Luna with \texttt{low} reasoning effort as a held-out judge on Physics-IQ Verified~\citep{DBLP:journals/corr/abs-2606-18943}, excluded from reward optimization. For each five-second continuation, the judge receives the best-practice prompt, conditioning image, and four $3\times5$ comparison grids. These grids align 20 uniformly sampled frames from the generated video and two real takes of the same scene and viewpoint.

Using category-specific criteria, the judge assigns integer scores from 1 to 5 for reference event fidelity $e_i$, trajectory fidelity $t_i$, physical plausibility $p_i$, temporal coherence $c_i$, and entity/scene consistency $s_i$. The real takes define the expected event progression while allowing natural variation. We retain one valid judgment per video, with reasons and frame evidence for each rating, and compute
\begin{equation}
\label{eq:vlm-judge-score}
\begin{aligned}
q_i&=25\!\left[\min\!\left(\frac{e_i+t_i}{2},\frac{2p_i+c_i+s_i}{4}\right)-1\right],
\\
J_{\mathrm{VLM}}&=\frac{1}{198}\sum_{i=1}^{198}q_i.
\end{aligned}
\end{equation}
We report $J_{\mathrm{VLM}}\in[0,100]$ in Table~\ref{tab:physics-iq-verified}, averaging over 198 videos from 66 scenes and three viewpoints.

For our evaluations on RBench~\citep{DBLP:journals/corr/abs-2601-15282}, we follow the released evaluation code and prompts, using the same GPT-5 snapshot (\texttt{gpt-5-2025-08-07}) and API seed (\texttt{2026}) as the official implementation. All models evaluated by us use the same 650 image--prompt pairs: 250 across five task categories (50 per category) and 400 across four embodiments (100 per embodiment).

\paragraph{Evaluation settings.}
Tables~\ref{tab:physics-iq-generation-settings} and~\ref{tab:rbench-generation-settings} report generation settings only for models evaluated by us on Physics-IQ Verified and RBench, respectively. FPS denotes the video frame rate, independently of the number of frames sampled by the evaluator. Cosmos3-Nano, Cosmos3-Nano + SFT, and CoDance use identical video-generation settings within each benchmark, generating five-second videos at $1280\times720$ resolution and 24 FPS, with classifier-free guidance of 6.0 and a flow shift of 10.0. Each video contains 121 frames, including the initial conditioning frame. For these three configurations, we use 35 denoising steps when evaluating on Physics-IQ Verified and 50 on RBench.

\begin{table*}[t]
    \centering
    \caption{Video-generation settings for models evaluated by us on Physics-IQ Verified. Duration denotes the five-second evaluated continuation.}
    \label{tab:physics-iq-generation-settings}
    \footnotesize
    \setlength{\tabcolsep}{3pt}
    \renewcommand{\arraystretch}{1.02}
    \begin{tabularx}{\linewidth}{@{}>{\raggedright\arraybackslash}Xcccccc@{}}
        \toprule
        \textbf{Model} & \textbf{Resolution} & \textbf{FPS} & \textbf{Duration} & \textbf{Steps} & \textbf{Guidance} & \textbf{Shift} \\
        \midrule
        \rowcolor{categoryblue}
        \multicolumn{7}{@{}l}{\textbf{\textit{Commercial}}} \\
        MiniMax-H3 Max & $1344\times768$ & 24 & 5 s & -- & -- & -- \\
        Seedance 2.0 & $1280\times720$ & 24 & 5 s & -- & -- & -- \\
        \midrule
        \rowcolor{categoryblue}
        \multicolumn{7}{@{}l}{\textbf{\textit{Open-source}}} \\
        Wan2.2-I2V-A14B & $1280\times720$ & 16 & 5 s & 35 & 3.5 & 5.0 \\
        HunyuanVideo 1.5 & $848\times480$ & 24 & 5 s & 35 & 6.0 & 5.0 \\
        Cosmos3-Super I2V & $832\times480$ & 24 & 5 s & 35 & 6.0 & 10.0 \\
        \midrule
        \rowcolor{categoryblue}
        \multicolumn{7}{@{}l}{\textbf{\textit{Physics-aware methods}}} \\
        PhiZero & $896\times512$ & 8 & 5 s & 50 & -- & 5.0 \\
        \midrule
        \rowcolor{categoryblue}
        \multicolumn{7}{@{}l}{\textbf{\textit{Ours and base model}}} \\
        Cosmos3-Nano & $1280\times720$ & 24 & 5 s & 35 & 6.0 & 10.0 \\
        Cosmos3-Nano + SFT & $1280\times720$ & 24 & 5 s & 35 & 6.0 & 10.0 \\
        \rowcolor{ourspurple}
        \textbf{CoDance} & $1280\times720$ & 24 & 5 s & 35 & 6.0 & 10.0 \\
        \bottomrule
    \end{tabularx}
    \vspace{3pt}
    \begin{minipage}{\linewidth}
        \scriptsize
        Resolution is given as width $\times$ height. Here, s denotes seconds and -- indicates an unreported or inapplicable value. Wan2.2-I2V-A14B uses 16 FPS and PhiZero uses 8 FPS, while the other listed models use 24 FPS. PhiZero pads 33 native frames to 40 scoring frames. The three Cosmos3-Nano-based configurations generate 121 frames and yield 120 scoring frames after removing the conditioning frame.
    \end{minipage}
\end{table*}

\begin{table*}[t]
    \centering
    \caption{Video-generation settings for models evaluated by us on RBench.}
    \label{tab:rbench-generation-settings}
    \footnotesize
    \setlength{\tabcolsep}{3pt}
    \renewcommand{\arraystretch}{1.02}
    \begin{tabularx}{\linewidth}{@{}>{\raggedright\arraybackslash}Xcccccc@{}}
        \toprule
        \textbf{Model} & \textbf{Resolution} & \textbf{FPS} & \textbf{Duration} & \textbf{Steps} & \textbf{Guidance} & \textbf{Shift} \\
        \midrule
        \rowcolor{categoryblue}
        \multicolumn{7}{@{}l}{\textbf{\textit{Commercial}}} \\
        Seedance 1.5 & $1280\times720$ & 24 & 5 s & -- & -- & -- \\
        Seedance 2.0 & -- & -- & -- & -- & -- & -- \\
        \midrule
        \rowcolor{categoryblue}
        \multicolumn{7}{@{}l}{\textbf{\textit{Ours and base model}}} \\
        Cosmos3-Nano & $1280\times720$ & 24 & 5 s & 50 & 6.0 & 10.0 \\
        Cosmos3-Nano + SFT & $1280\times720$ & 24 & 5 s & 50 & 6.0 & 10.0 \\
        \rowcolor{ourspurple}
        \textbf{CoDance} & $1280\times720$ & 24 & 5 s & 50 & 6.0 & 10.0 \\
        \bottomrule
    \end{tabularx}
    \vspace{3pt}
    \begin{minipage}{\linewidth}
        \scriptsize
        Resolution is given as width $\times$ height. Here, s denotes seconds and -- indicates an unreported or inapplicable value. The three Cosmos3-Nano-based configurations generate five-second videos at 24 FPS with 121 frames, including the initial conditioning frame.
    \end{minipage}
\end{table*}

For WorldArena, we follow the official action-planning protocol on the RoboTwin 2.0 Clean-50 data, using 40 training episodes and ten held-out test episodes per task. We keep both the reasoner and generator frozen and train a task-specific inverse dynamics model (IDM) with the official architecture and training procedure. During control, the reasoner generates a prompt rewrite that conditions the generator. The IDM reads the generator's features to predict an action chunk; no future video is generated. The controller refreshes its observation after executing the chunk. RoboTwin 2.0 demonstrations are used to train the IDM, but are excluded from CoDance's SFT and RL data. We evaluate \texttt{adjust\_bottle} and \texttt{click\_bell}, repeating each of the ten test initial conditions ten times for 100 rollouts per task. We use the official task-success criteria and report the success rate for each task and their arithmetic mean.

\section{Additional Experiments}
\label{app:additional-experiments}

\subsection{Full VLM-as-Judge Results}
\label{app:vlm-judge-breakdown}

The held-out judge evaluates five aspects of each Physics-IQ Verified continuation (Table~\ref{tab:vlm-judge-breakdown}). CoDance leads on all five. Its overall score is 61.39, compared with 31.69 for Cosmos3-Nano and 41.89 for Seedance 2 Pro. The largest gain over Cosmos3-Nano is in temporal coherence (45.45 to 77.68).

\subsection{Training Strategies and Conditioning Features}
\label{app:training-component-ablation}

Table~\ref{tab:ablation-study} compares coupled optimization with reasoner-only, generator-only, and sequential reinforcement learning. All RL configurations start from the same component-wise SFT checkpoints and use both the physical-fidelity reward and the geometric trajectory reward introduced in Section~\ref{sec:rewards}. The total numbers of sampled video rollouts and optimizer updates are matched across configurations. The sequential baseline's budget includes both training stages. For both GRPO and DanceGRPO, including their respective stages in the sequential baseline, we use GDPO's reward-decoupled normalization. Each reward is normalized separately before aggregation, and the combined signal is then normalized over the batch. CoDanceRL applies this principle at both the prompt and video levels through the hierarchical construction in Section~\ref{sec:normalization}.

Reasoner-only GRPO updates the reasoner while keeping the SFT generator frozen, reaching an IQ-Score of 42.0. Generator-only DanceGRPO updates the generator while keeping the SFT reasoner frozen, reaching 39.3. These configurations improve over the SFT initialization (37.6) by 4.4 and 1.7 points, respectively. The sequential GRPO+DanceGRPO baseline first trains the reasoner with the generator frozen, then freezes the resulting reasoner and trains the generator, following the staged pattern of VANS~\citep{DBLP:conf/cvpr/ChengHTL26}. It reaches 44.2, exceeding both single-component configurations, although the reasoner receives no further updates as the generator adapts in the second stage.

CoDanceRL keeps both components trainable throughout the same run. Each iteration updates the reasoner and generator using their respective advantages from a shared nested rollout. The reasoner continues to receive feedback as the generator adapts.

\filbreak
Coupled training reaches an IQ-Score of 48.3, exceeding sequential training by 4.1 points, reasoner-only training by 6.3, and generator-only training by 9.0. CoDance leads on all three IoU metrics in Table~\ref{tab:ablation-study}. The sequential baseline has the highest reverse-normalized MSE score.

\paragraph{Conditioning-encoder comparison.}
The VLM rewrites the prompt and encodes the conditioning features used by the generator. Table~\ref{tab:ablation-overview}(a) holds the rewrite and generator checkpoint fixed within each training strategy. Replacing the base encoder with the encoder learned under that strategy isolates the effect of the conditioning features.

\filbreak
With the CoDance generator, using the trained encoder raises IQ-Score from 45.4 to 48.3. The gains are 0.5 points for reasoner-only GRPO and 0.1 for sequential GRPO+DanceGRPO. For SFT, the score falls from 38.4 to 37.6. CoDance thus shows the largest encoder gain with the rewrite held fixed.

\paragraph{Inference-time component swaps.}
We also pair complete reasoner and generator checkpoints at inference time (Section~\ref{app:feature-generator-compatibility}). Fixing the base generator lets us compare reasoners, and fixing the base reasoner lets us compare generators. The mixed SFT and CoDance pairs test how each component performs with a partner from a different training stage.

\twocolumn[{%
    \begin{minipage}{\textwidth}
        \centering
        \captionsetup{type=table,position=top,skip=6pt}
        \caption{Held-out VLM judge scores on Physics-IQ Verified (0--100, higher is better). The final column reports the overall per-video score. Bold marks the best result in each column.}
        \label{tab:vlm-judge-breakdown}
        {\footnotesize
        \setlength{\tabcolsep}{3pt}
        \renewcommand{\arraystretch}{1.12}
        \begin{tabularx}{\linewidth}{@{}l*{6}{>{\raggedleft\arraybackslash}X}@{}}
            \toprule
            \textbf{Model} & \textbf{\shortstack{Reference event\\fidelity}} & \textbf{\shortstack{Trajectory\\fidelity}} & \textbf{\shortstack{Physical\\plausibility}} & \textbf{\shortstack{Temporal\\coherence}} & \textbf{\shortstack{Entity/scene\\consistency}} & \textbf{\shortstack{VLM as\\Judge}} \\
            \midrule
            HunyuanVideo-1.5 & 34.34 & 23.48 & 30.68 & 41.79 & 45.33 & 26.70 \\
            Cosmos3-Nano     & 37.25 & 27.53 & 40.40 & 45.45 & 64.27 & 31.69 \\
            Wan2.2           & 40.40 & 28.91 & 37.25 & 45.45 & 59.72 & 33.18 \\
            PhiZero          & 38.51 & 30.30 & 46.97 & 49.62 & 73.11 & 33.90 \\
            Cosmos3-Super I2V & 42.68 & 32.70 & 42.30 & 51.89 & 63.64 & 36.40 \\
            Seedance 2 Pro   & 47.22 & 37.25 & 51.14 & 55.68 & 73.61 & 41.89 \\
            \rowcolor{ourspurple}
            \textbf{CoDance} & \textbf{69.22} & \textbf{56.34} & \textbf{66.19} & \textbf{77.68} & \textbf{82.98} & \textbf{61.39} \\
            \bottomrule
        \end{tabularx}\par}
        \bigskip
        \captionsetup{type=table,position=top,skip=6pt}
        \caption{Training strategies on Physics-IQ Verified. All RL runs share SFT initialization and matched rollout and update budgets. All metrics, including reverse-normalized MSE, favor higher values. Bold and underlining mark the best and second-best results.}
        \label{tab:ablation-study}
        {\small
        \setlength{\tabcolsep}{4pt}
        \renewcommand{\arraystretch}{1.12}
        \begin{tabularx}{\linewidth}{@{}l*{5}{>{\raggedleft\arraybackslash}X}@{}}
            \toprule
            \textbf{Model} & \textbf{S-IoU} & \textbf{ST-IoU} & \textbf{W-S-IoU} & \textbf{MSE} & \textbf{IQ-Score} \\
            \midrule
            Cosmos3-Nano (base) & 45.2 & 26.5 & 31.3 & 36.3 & 34.8 \\
            SFT & 44.3 & 38.0 & 32.5 & 35.6 & 37.6 \\
            GRPO (w/ SFT)~\citep{DBLP:journals/corr/abs-2402-03300} & 50.4 & 42.6 & 38.0 & \underline{36.9} & 42.0 \\
            DanceGRPO (w/ SFT)~\citep{DBLP:journals/corr/abs-2505-07818} & 51.2 & 34.2 & 37.9 & 33.7 & 39.3 \\
            GRPO+DanceGRPO (w/ SFT) & \underline{55.0} & \underline{43.1} & \underline{41.1} & \textbf{37.6} & \underline{44.2} \\
            \midrule
            \rowcolor{ourspurple}
            \textbf{CoDance (joint training)} & \textbf{60.2} & \textbf{50.0} & \textbf{48.3} & 34.5 & \textbf{48.3} \\
            \bottomrule
        \end{tabularx}\par}
        \bigskip
        \begin{minipage}[t]{0.48\linewidth}
            \vspace{0pt}
            \centering
            \captionsetup{type=table,position=top,skip=6pt}
            \caption{IQ-Score on Physics-IQ Verified~\citep{DBLP:journals/corr/abs-2606-18943} for reasoner and generator checkpoints paired at inference time.}
            \label{tab:component-composition}
            \footnotesize
            \setlength{\tabcolsep}{5pt}
            \begin{tabularx}{\linewidth}{@{}>{\raggedright\arraybackslash}X>{\raggedright\arraybackslash}Xr@{}}
                \toprule
                \textbf{Reasoner} & \textbf{Generator} & \textbf{IQ-Score} \\
                \midrule
                Base    & Base    & 34.8 \\
                CoDance & Base    & 45.4 \\
                Base    & CoDance & 37.6 \\
                CoDance & SFT     & \underline{46.0} \\
                SFT     & CoDance & 37.0 \\
                \midrule
                \rowcolor{ourspurple}
                \textbf{CoDance} & \textbf{CoDance} & \textbf{48.3} \\
                \bottomrule
            \end{tabularx}
        \end{minipage}\hfill
        \begin{minipage}[t]{0.48\linewidth}
            \vspace{0pt}
            \centering
            \captionsetup{type=table,position=top,skip=6pt}
            \caption{Closed-loop success rates (\%) for the RoboTwin 2.0 tasks in Sec.~\ref{app:additional-robotwin-tasks}. Avg.\ is the unweighted mean.}
            \label{tab:additional-robotwin-tasks}
            \footnotesize
            \setlength{\tabcolsep}{3pt}
            \begin{tabularx}{\linewidth}{@{}>{\raggedright\arraybackslash}Xrrrr@{}}
                \toprule
                \textbf{Model} & \textbf{Lift Pot} & \textbf{Press Stapler} & \textbf{Shake Bottle} & \textbf{Avg.} \\
                \midrule
                Cosmos3-Nano & 13.0 & 29.0 & 5.0 & 15.7 \\
                \rowcolor{ourspurple}
                \textbf{CoDance} & \textbf{37.0} & \textbf{41.0} & \textbf{49.0} & \textbf{42.3} \\
                \bottomrule
            \end{tabularx}
        \end{minipage}\par
        \vspace{12pt}
    \end{minipage}
}]

\subsection{Component Compatibility\texorpdfstring{\strut}{}}
\label{app:feature-generator-compatibility}

We recombine the base, SFT, and CoDance checkpoints without further training (Table~\ref{tab:component-composition}). With the base generator fixed, switching to the CoDance reasoner raises IQ-Score from 34.8 to 45.4. With the base reasoner fixed, switching to the CoDance generator raises it to 37.6.

The CoDance reasoner scores 46.0 with the SFT generator. The reverse pairing scores 37.0. The jointly trained pair reaches 48.3, exceeding these mixed pairs by 2.3 and 11.3 points, respectively.

\newpage
\subsection{Reward and Normalization Ablations}
\label{app:normalization-analysis}

\begin{figure}[!ht]
    \centering
    \includegraphics[width=0.82\linewidth]{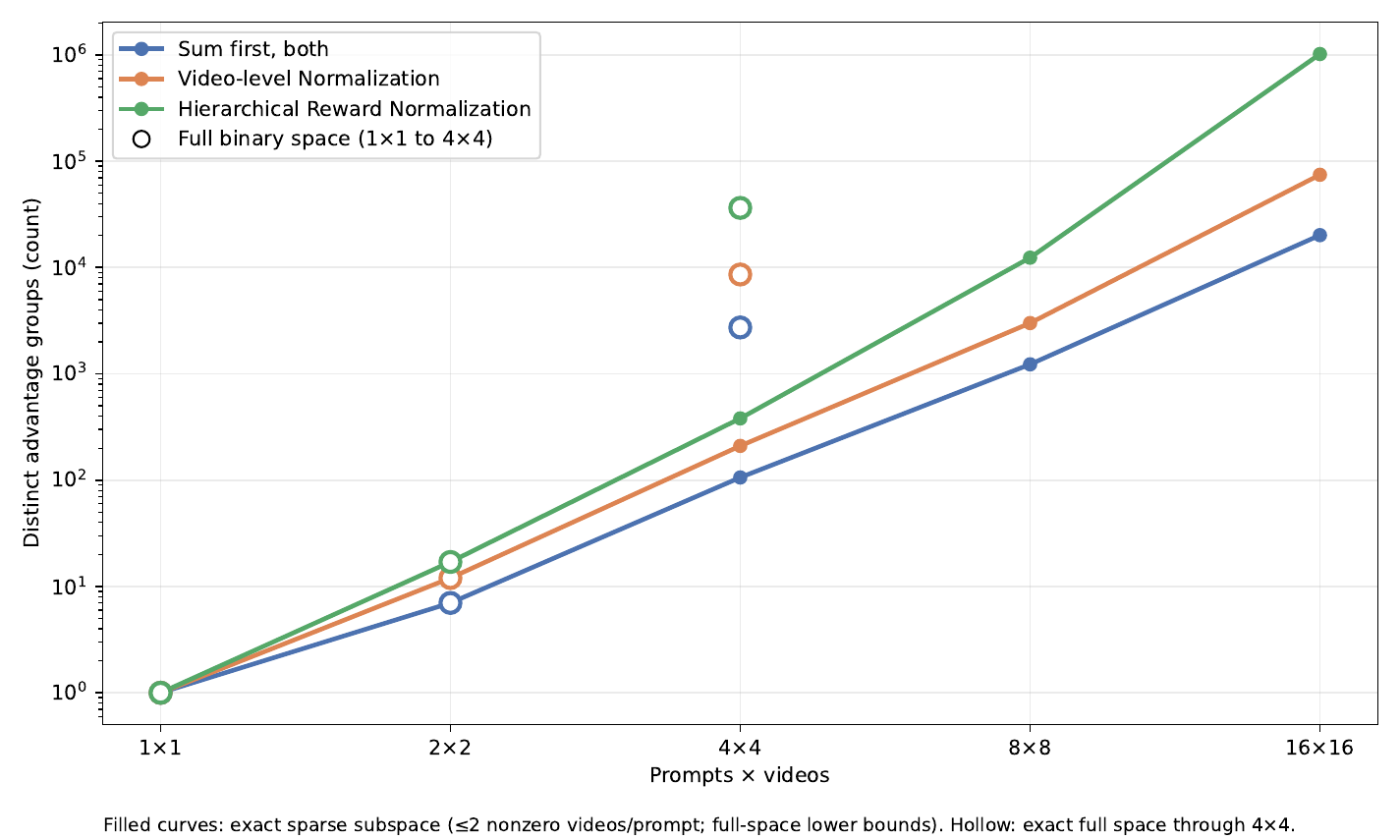}
    \caption{\textbf{Distinct advantage groups under three normalization rules.} Hollow markers give exact full-space counts through $4\times4$. Filled markers count a sparse subspace with at most two videos receiving a nonzero reward per prompt. These counts are lower bounds for the full space.}
    \label{fig:advantage-group-counts}
\end{figure}

Table~\ref{tab:reward-normalization-ablation} compares reward terms and normalization levels on Physics-IQ Verified. Figure~\ref{fig:advantage-group-counts} counts the distinct advantage groups produced by three normalization rules.

\noindent\textbf{Reward terms.} Before reward-decoupled normalization, the physical-fidelity reward alone scores 42.8 and the geometric trajectory reward alone scores 40.1. Combining them yields 41.3.

\smallskip
\noindent\textbf{Normalization levels.} With both rewards enabled, video-level normalization raises IQ-Score from 41.3 to 43.9 and prompt-level normalization raises it to 43.3. Using both levels reaches 48.3.

\onecolumn
\begin{table}[!ht]
    \centering
    \caption{Reward terms and normalization levels on Physics-IQ Verified. Checkmarks (\cmark) mark enabled components. Higher IQ-Score is better.}
    \label{tab:reward-normalization-ablation}
    \small
    \setlength{\tabcolsep}{4pt}
    \renewcommand{\arraystretch}{1.05}
    \begin{tabularx}{\linewidth}{@{}>{\raggedright\arraybackslash}X*{4}{>{\centering\arraybackslash}p{43pt}}r@{}}
        \toprule
        \multirow{2}{*}{\textbf{Configuration}} & \multicolumn{2}{c}{\textbf{Reward}} & \multicolumn{2}{c}{\shortstack{\textbf{Hierarchical Reward}\\\textbf{Normalization}}} & \multirow{2}{*}{\textbf{IQ-Score}$\uparrow$} \\
        \cmidrule(lr){2-3}\cmidrule(lr){4-5}
        & \shortstack{Physical\\fidelity} & \shortstack{Geometric\\trajectory} & \shortstack{Video\\level} & \shortstack{Prompt\\level} & \\
        \midrule
        Physical-fidelity reward   & \cmark & \xmark & \xmark & \xmark & 42.8 \\
        Geometric trajectory reward & \xmark & \cmark & \xmark & \xmark & 40.1 \\
        Combined rewards           & \cmark & \cmark & \xmark & \xmark & 41.3 \\
        Video-level normalization  & \cmark & \cmark & \cmark & \xmark & \underline{43.9} \\
        Prompt-level normalization & \cmark & \cmark & \xmark & \cmark & 43.3 \\
        \midrule
        \rowcolor{ourspurple}
        \textbf{CoDance (full)}     & \cmark & \cmark & \cmark & \cmark & \textbf{48.3} \\
        \bottomrule
    \end{tabularx}
\end{table}

\paragraph{Distinct advantage groups.}
Figure~\ref{fig:advantage-group-counts} compares summing rewards before normalization at both levels, normalizing each reward only at the video level, and normalizing each reward at both levels. We increase the numbers of prompts and videos per prompt together from $1\times1$ to $16\times16$. Beyond $1\times1$, hierarchical normalization yields the most distinct groups in both enumerations. These counts describe the training signals. Table~\ref{tab:reward-normalization-ablation} reports video quality.

\subsection{Additional RoboTwin 2.0 Tasks}
\label{app:additional-robotwin-tasks}

We extend the closed-loop evaluation in Section~\ref{sec:embodied-video-generation} to \texttt{lift\_pot}, \texttt{press\_stapler}, and \texttt{shake\_bottle}. We train one IDM per task on RoboTwin 2.0 demonstrations (Section~\ref{sec:experimental-setup}). CoDance's SFT and RL data exclude RoboTwin 2.0.

CoDance improves on all three tasks (Table~\ref{tab:additional-robotwin-tasks}). Mean success rises from 15.7\% to 42.3\%, with the largest gain on Shake Bottle (5.0\% to 49.0\%).

\begin{figure}[p]
    \centering
    \captionsetup{skip=7pt}
    \includegraphics[width=\linewidth]{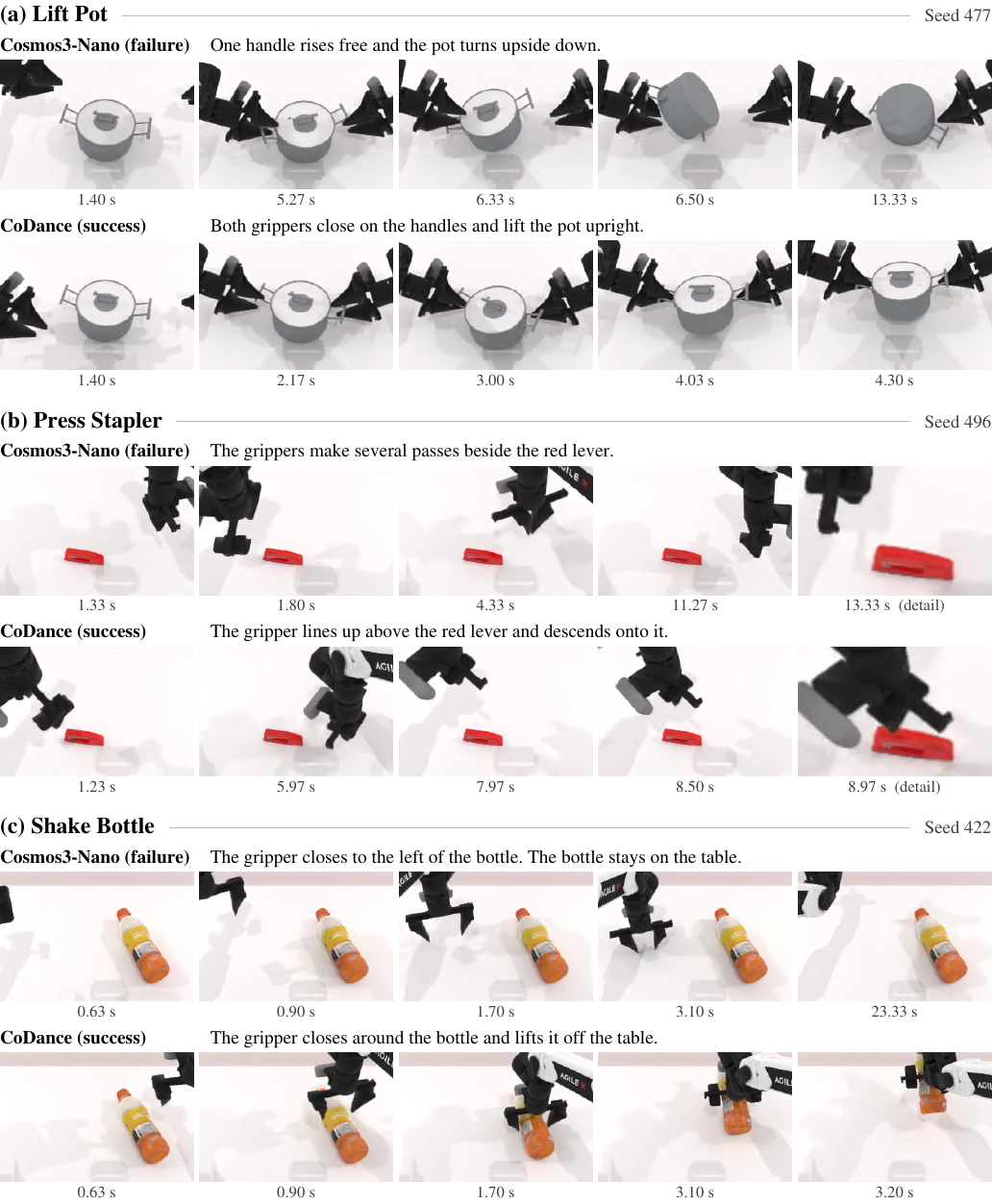}
    \caption{\textbf{RoboTwin 2.0 rollouts for Table~\ref{tab:additional-robotwin-tasks}.} Each pair starts from the same state. Timestamps give elapsed video time. The last column shows the final frames, with matching close-ups of the stapler lever in (b).}
    \label{fig:robotwin-rollouts}
\end{figure}

Figure~\ref{fig:robotwin-rollouts} follows one pair of rollouts per task. On Lift Pot, one handle rises clear of the Cosmos3-Nano gripper at 6.33\,s and the pot turns onto its side by 6.50\,s. CoDance holds the pot upright by both handles at 4.30\,s.

\clearpage

\twocolumn
\makeatletter
\setlength{\@dblfpsep}{6pt plus 0.4fil}
\makeatother
\newcommand{\physicsiqgrid}[1]{%
    \centering
    \includegraphics[trim=0bp 5bp 0bp 3bp,clip,width=0.95\linewidth]{visualQ/#1.pdf}%
}

\begin{figure*}[p]
    \section{Additional Qualitative Results}
    \label{app:more-visualization-results}
    \subsection{Physics-IQ Verified}
    {\normalsize\noindent Figures~\ref{fig:additional-physics-iq-rotating-cardstock}--\ref{fig:additional-physics-iq-fabric-box} compare Cosmos3-Nano, Seedance 2.0, and CoDance on six Physics-IQ Verified examples. Frames run left to right in temporal order, with the evaluated prompt below each grid.\par}
    \medskip
    \physicsiqgrid{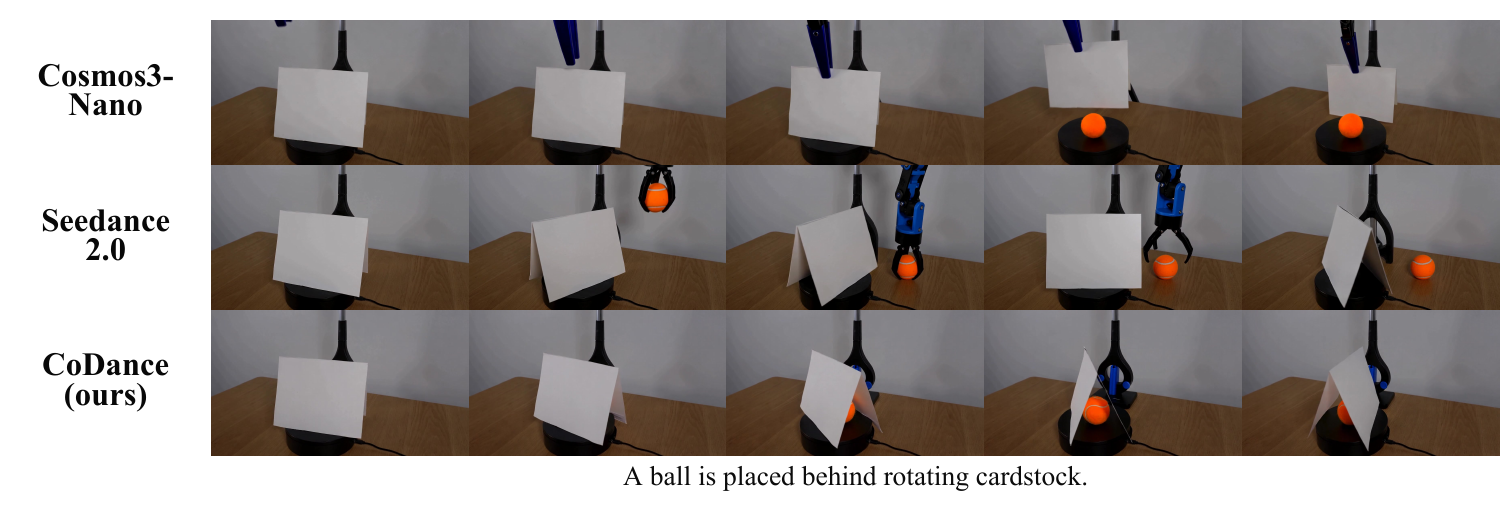}
    \caption{\textbf{Additional Physics-IQ Verified comparison of a ball behind rotating cardstock.}}
    \label{fig:additional-physics-iq-rotating-cardstock}
\end{figure*}

\begin{figure*}[p]
    \physicsiqgrid{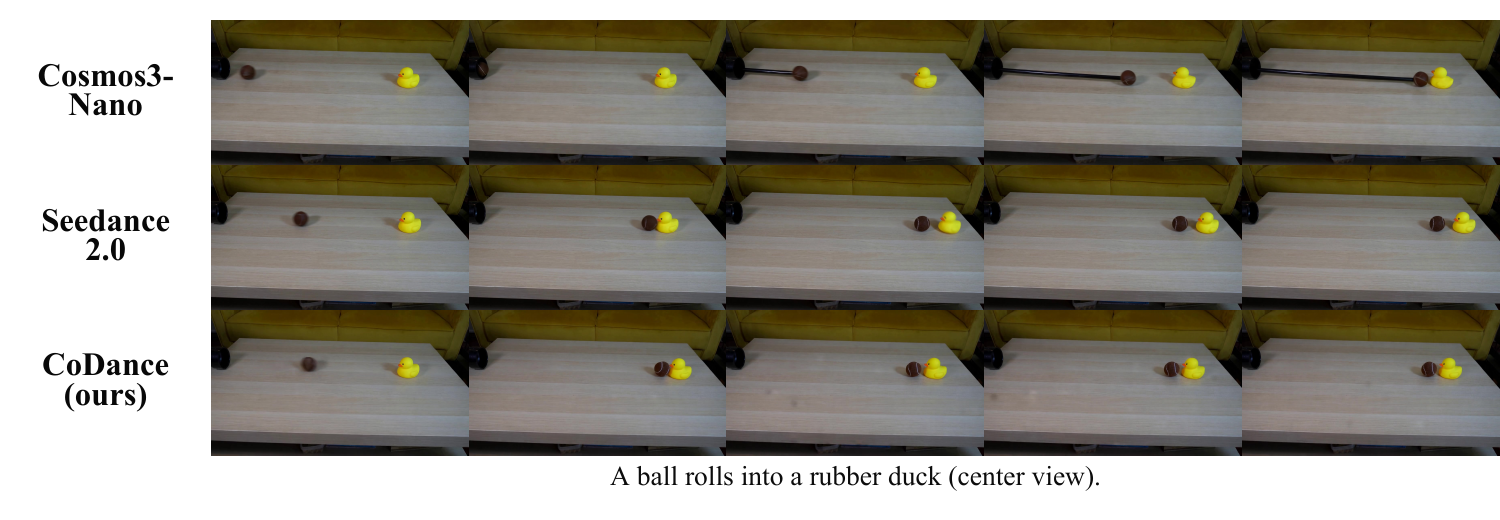}
    \caption{\textbf{Additional Physics-IQ Verified comparison of a ball colliding with a rubber duck (center view).}}
    \label{fig:additional-physics-iq-duck-center}
\end{figure*}

\begin{figure*}[p]
    \physicsiqgrid{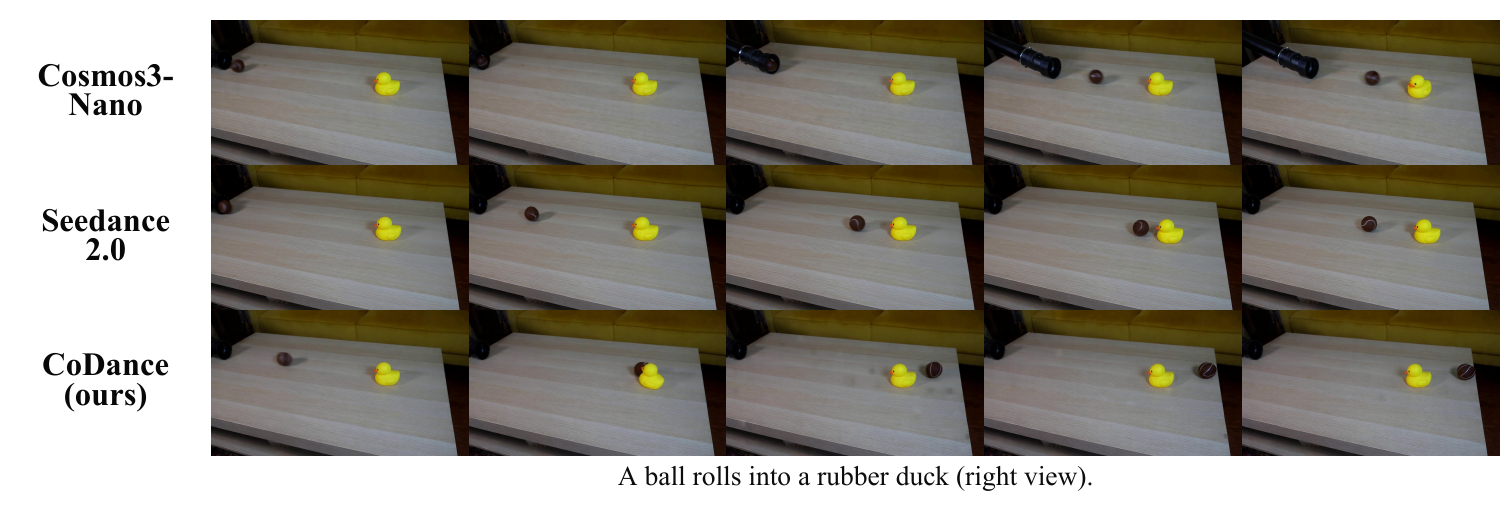}
    \caption{\textbf{Additional Physics-IQ Verified comparison of a ball colliding with a rubber duck (right view).}}
    \label{fig:additional-physics-iq-duck-right}
\end{figure*}

\begin{figure*}[p]
    \physicsiqgrid{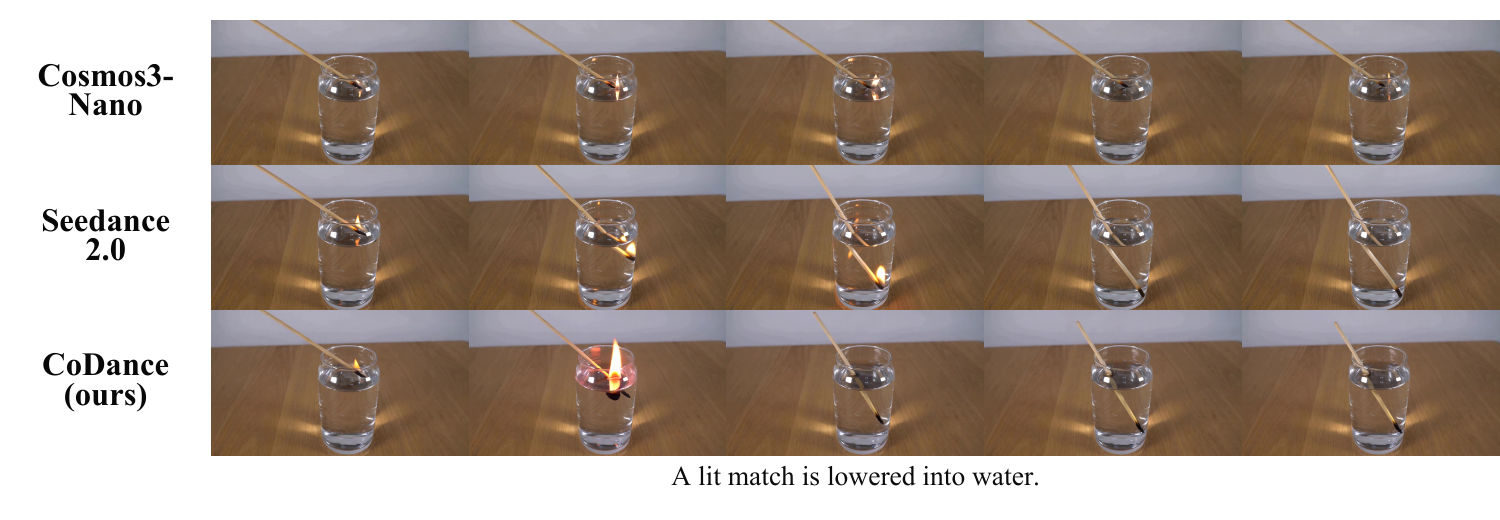}
    \caption{\textbf{Additional Physics-IQ Verified comparison of a lit match lowered into water.}}
    \label{fig:additional-physics-iq-match}
\end{figure*}

\begin{figure*}[p]
    \physicsiqgrid{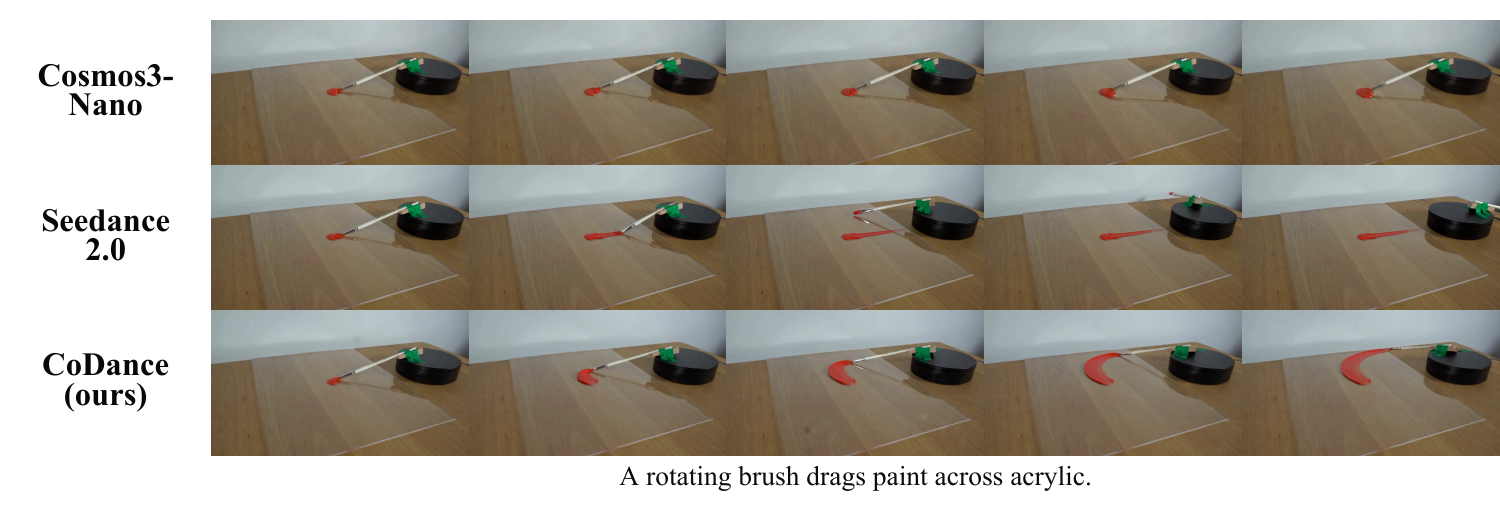}
    \caption{\textbf{Additional Physics-IQ Verified comparison of a rotating brush dragging paint across acrylic.}}
    \label{fig:additional-physics-iq-paint}
\end{figure*}

\begin{figure*}[p]
    \physicsiqgrid{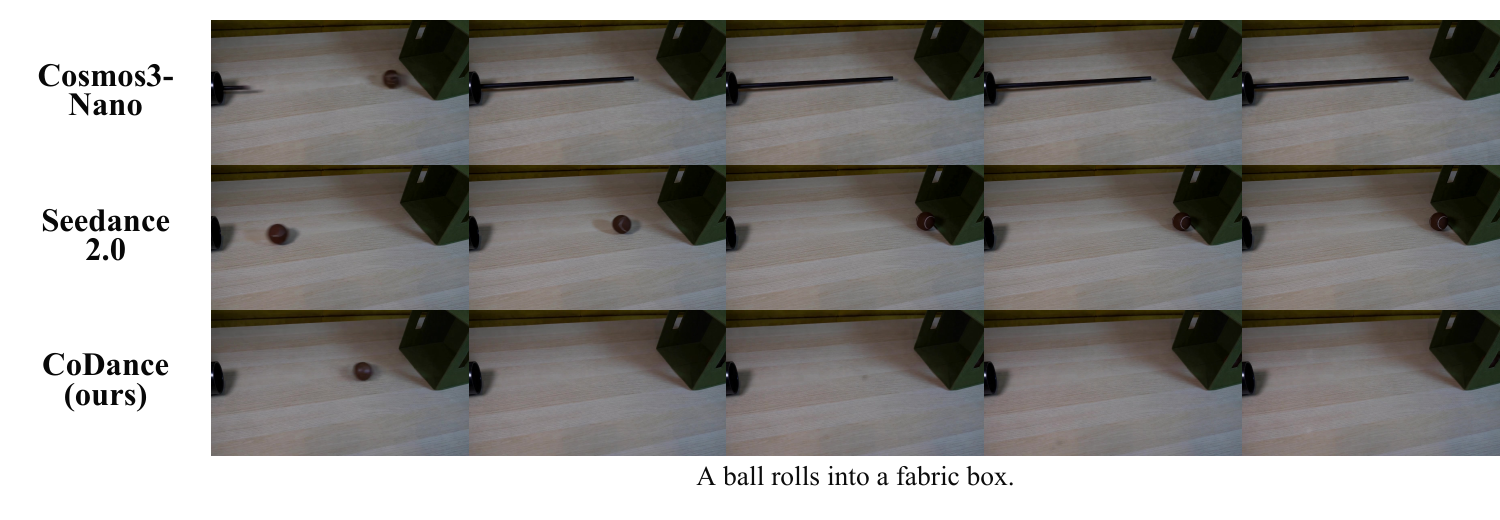}
    \caption{\textbf{Additional Physics-IQ Verified comparison of a ball rolling into a fabric box.}}
    \label{fig:additional-physics-iq-fabric-box}
\end{figure*}

\clearpage
\newcommand{\rbenchgrid}[1]{%
    \centering
    \captionsetup{skip=3pt}%
    \includegraphics[width=0.98\linewidth]{visualQ/seedance2/#1.pdf}%
}

\begin{figure*}[p]
    \subsection{RBench}
    {\normalsize\noindent Figures~\ref{fig:additional-rbench-bread-toaster}--\ref{fig:additional-rbench-block-cup} compare Cosmos3-Nano, Seedance 2.0, and CoDance (top to bottom) on seven RBench tasks. Each caption gives the requested action. Frames run left to right, with Seedance 2.0 sampled every 1.25\,s from 0 to 5\,s.\par}
    \medskip
    \rbenchgrid{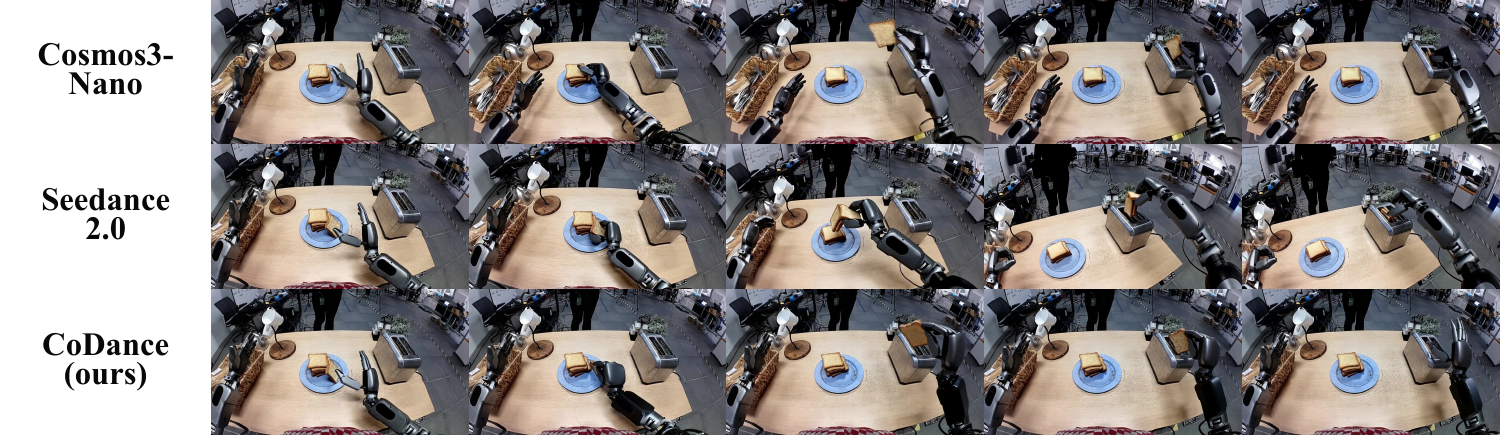}
    \caption{\textbf{The humanoid robot inserts a slice of bread into the toaster.}}
    \label{fig:additional-rbench-bread-toaster}
\end{figure*}

\begin{figure*}[p]
    \rbenchgrid{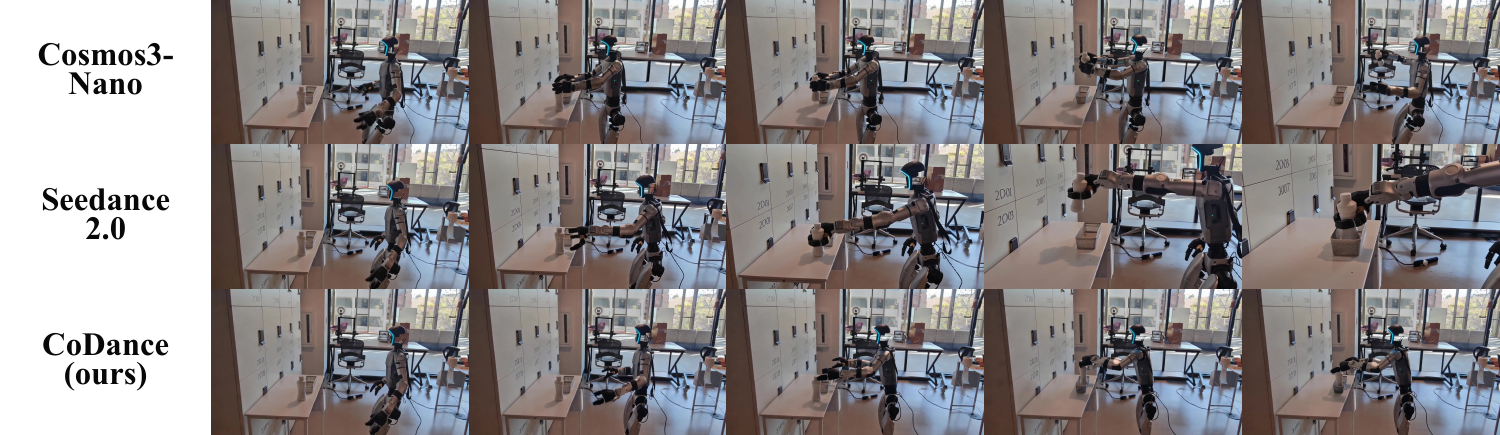}
    \caption{\textbf{The humanoid robot places the bottle into the mesh basket.}}
    \label{fig:additional-rbench-bottle-basket}
\end{figure*}

\begin{figure*}[p]
    \rbenchgrid{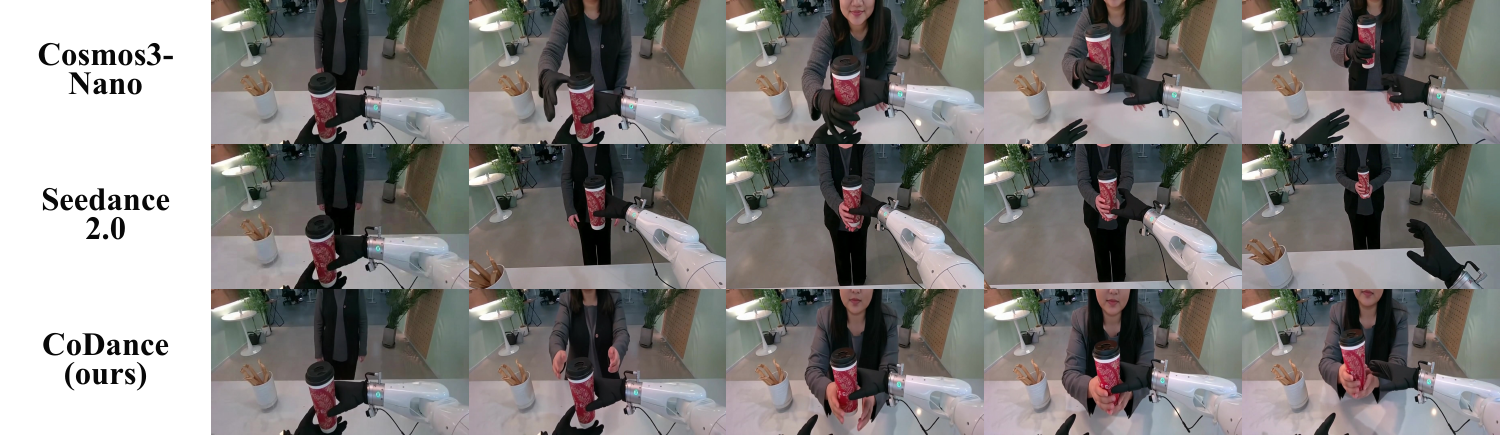}
    \caption{\textbf{The robotic hand passes the red coffee cup to the woman.}}
    \label{fig:additional-rbench-cup-handoff}
\end{figure*}

\begin{figure*}[p]
    \rbenchgrid{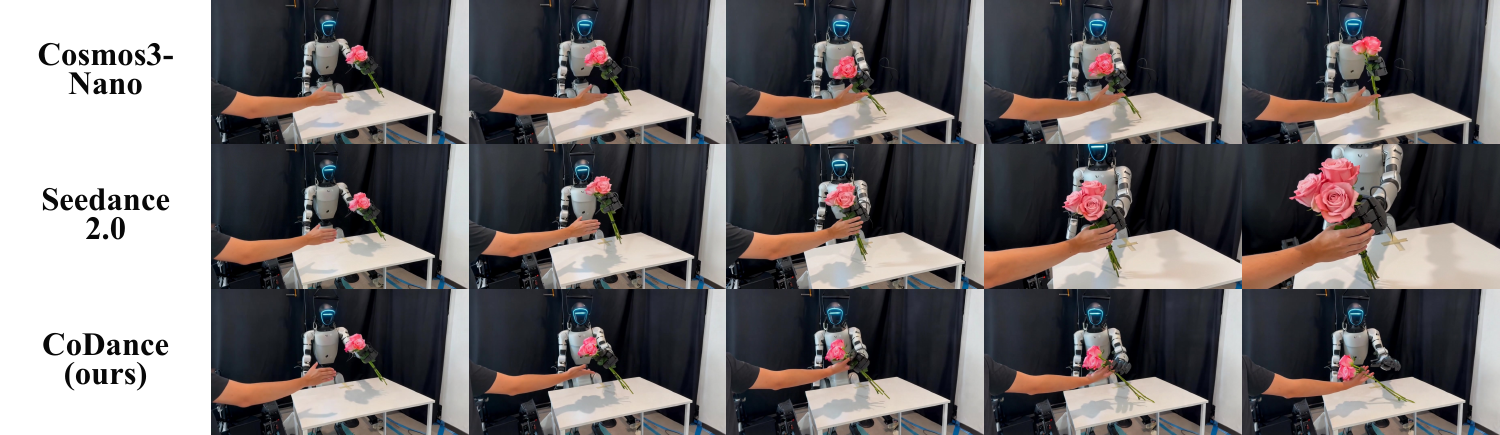}
    \caption{\textbf{The humanoid robot hands the flower bouquet to the person.}}
    \label{fig:additional-rbench-bouquet-handoff}
\end{figure*}

\begin{figure*}[p]
    \rbenchgrid{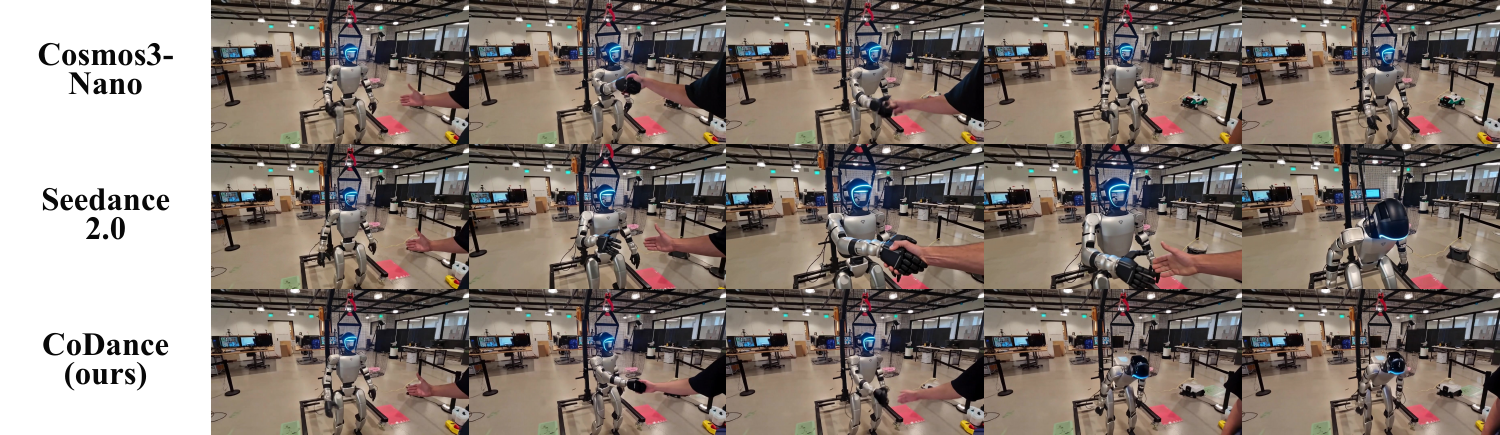}
    \caption{\textbf{The humanoid robot shakes hands, then bows politely.}}
    \label{fig:additional-rbench-handshake-bow}
\end{figure*}

\begin{figure*}[p]
    \rbenchgrid{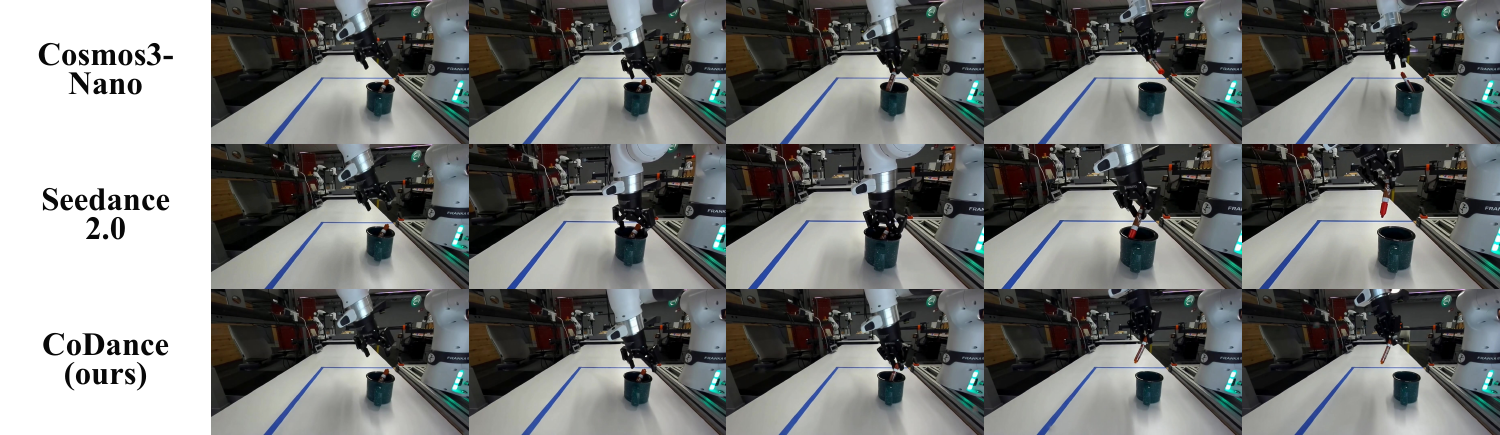}
    \caption{\textbf{The gripper removes the red marker from the green mug.}}
    \label{fig:additional-rbench-marker-mug}
\end{figure*}

\begin{figure*}[p]
    \rbenchgrid{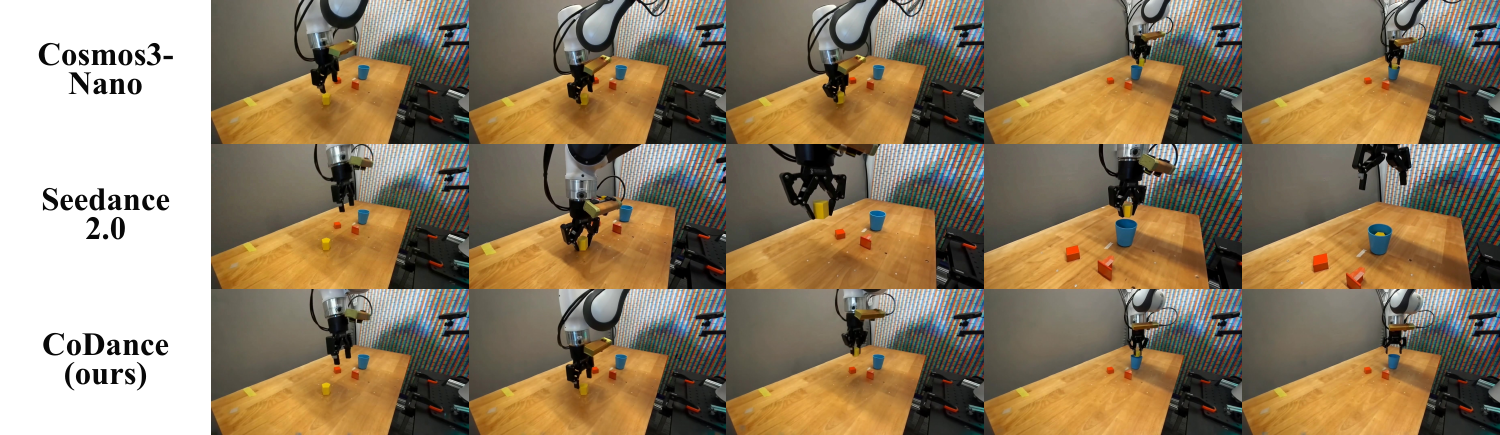}
    \caption{\textbf{The gripper places the yellow toy block into the blue cup.}}
    \label{fig:additional-rbench-block-cup}
\end{figure*}

\clearpage